\documentclass{article}

\usepackage[T1]{fontenc}
\usepackage{iclr2027_conference,times}

\usepackage{amsmath,amsfonts,bm}

\def\eqref#1{equation~\ref{#1}}

\def\1{\bm{1}}

\DeclareMathAlphabet{\mathsfit}{\encodingdefault}{\sfdefault}{m}{sl}
\SetMathAlphabet{\mathsfit}{bold}{\encodingdefault}{\sfdefault}{bx}{n}

\usepackage{amssymb}

\usepackage{url}
\usepackage{graphicx}
\usepackage{booktabs}
\usepackage{array}
\usepackage{capt-of}
\usepackage{placeins}
\usepackage{xcolor}
\usepackage{algorithm}
\usepackage{algpseudocode}
\usepackage{hyperref}
\definecolor{citered}{rgb}{0.878,0.243,0.243}
\hypersetup{colorlinks=true, citecolor=citered, urlcolor=citered, linkcolor=blue}
\usepackage{etoolbox}
\makeatletter
\patchcmd{\NAT@citex}
  {\@citea\NAT@hyper@{\NAT@nmfmt{\NAT@nm}\hyper@natlinkbreak{\NAT@aysep\NAT@spacechar}{\@citeb\@extra@b@citeb}\NAT@date}}
  {\@citea\NAT@nmfmt{\NAT@nm}\NAT@aysep\NAT@spacechar\NAT@hyper@{\NAT@date}}
  {}{\PackageWarning{main}{natbib year-only link patch failed}}
\makeatother

\makeatletter
\def\@maketitle{\vbox{\hsize\textwidth
{\centering\LARGE\sc \@title\par}
\vskip 0.1in
\begin{tabular}[t]{l}\rule{\z@}{24pt}\@author\end{tabular}%
\vskip 0.3in minus 0.1in}}
\makeatother

\newcommand{\tightcaption}{\par\nointerlineskip\setlength{\abovecaptionskip}{6pt}}

\title{ReplaySSM: Cache SSM Inputs, Not State}

\author{%
\begin{minipage}[t]{\dimexpr\textwidth-2\tabcolsep\relax}
\centering\large
Ze-Wei Liou\textsuperscript{1}, Harris Nover\textsuperscript{2},
Amey Naik\textsuperscript{2}, Andrii Skliar\textsuperscript{2}\\[2pt]
Igor Shovkun\textsuperscript{2}, Ido Shahaf\textsuperscript{2},
Tri Dao\textsuperscript{1}\\[8pt]
\normalsize
\textsuperscript{1}Princeton University\qquad\textsuperscript{2}NVIDIA\\[5pt]
{\footnotesize\ttfamily
\{zewei.liou, tridao\}@princeton.edu\\
\{hnover, ameyn, askliar, ishovkun, ishahaf\}@nvidia.com}
\end{minipage}}

\iclrfinalcopy
\hypersetup{
    pdftitle={ReplaySSM: Cache SSM Inputs, Not State},
    pdfauthor={Ze-Wei Liou, Harris Nover, Amey Naik, Andrii Skliar, Igor Shovkun, Ido Shahaf, Tri Dao}
}

\begin{document}

\addtocontents{toc}{\protect\setcounter{tocdepth}{-1}}

\maketitle
\fancyhead{}
\renewcommand{\headrulewidth}{0pt}

\begin{abstract}
State-space models (SSMs) maintain a fixed-size recurrent state, but reading and
writing it at every decoding step can bottleneck hybrid models. This eager state
management causes large state I/O with low arithmetic intensity, leaves
speculative decoding no cheap rollback, and feeds state quantization error into
subsequent recurrent updates. We introduce ReplaySSM, which unrolls the
recurrence to represent the state as a checkpoint and a buffer of recent
SSM inputs. This separates output computation from checkpoint updates. By
updating the checkpoint periodically, ReplaySSM avoids state writes on most
steps and roughly halves the dominant state memory traffic. Keeping recent inputs also
makes speculative rollback a buffer-pointer operation and enables output-only
verification without materializing intermediate states. The same separation
allows a low-precision checkpoint to serve output computation on most steps
while a higher-precision checkpoint serves recurrent updates, so rounding
error in the low-precision copy never enters later updates. We implement
ReplaySSM in vLLM for Mamba-2 and Gated DeltaNet. At serving batch size 256, it
accelerates standard decoding by $1.20$--$1.47\times$ over a tuned vLLM baseline
across four hybrid models, and delivers about $1.8\times$ the throughput of
vLLM's speculative decoding on the two 120B-class models. Separate state
precisions also reduce output KL divergence and downstream accuracy losses
compared with a shared low-precision checkpoint.
\par\noindent Code: \url{https://github.com/Johnny-Liou/ReplaySSM}.
\end{abstract}

\section{Introduction}
\label{sec:introduction}

Long chain-of-thought reasoning makes decoding a substantial cost in LLM
inference. Agent workloads stretch it further through repeated tool calls and
reasoning. Transformer attention keeps past keys and values in a key-value
(KV) cache
whose size and per-step memory traffic grow with sequence length, hurting both
throughput and tail latency. State-space models (SSMs), such as Mamba-2
\citep{dao2024transformers}, Gated DeltaNet (GDN; \citealp{yang2025gated}),
and Kimi Delta Attention (KDA; \citealp{team2025kimi}), instead summarize past
information into a fixed-size recurrent state. Their state storage and
per-step memory traffic remain constant with sequence length, but this
compression can degrade exact recall
\citep{jelassi2024repeat,waleffe2024empirical}. Hybrid architectures aim to
find the sweet spot in between by interleaving SSM and attention layers and
are widely adopted in production models, including Nemotron-3
\citep{blakeman2025nvidia}, Qwen3.5 through Qwen3.8
\citep{qwen35blog,qwen38blog}, Kimi-K3 \citep{team2026kimi}, and
GLM-5.3-Flash \citep{zai2026glm53flash}.

\begin{figure}[!t]
    \centering
    \includegraphics[width=0.98\linewidth]{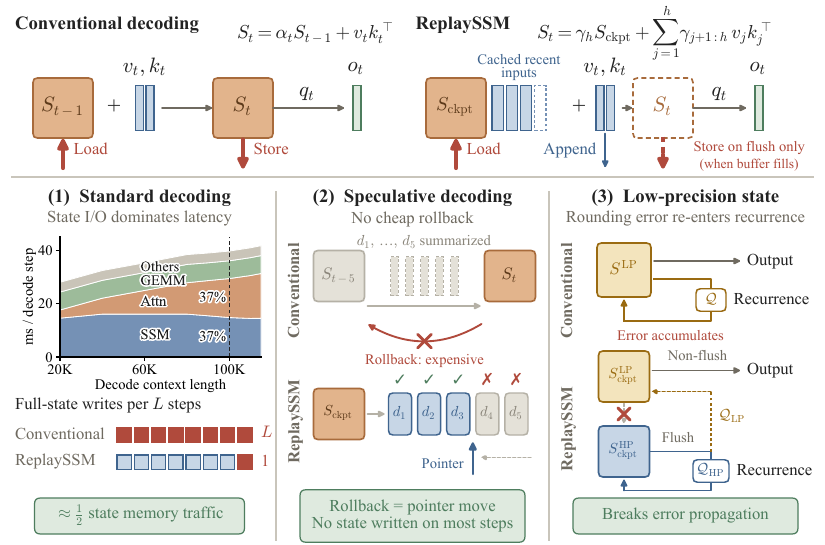}
    \caption{ReplaySSM overview. \textbf{Top:} conventional decoding loads and
    stores the full state at every step. ReplaySSM unrolls the recurrence to
    separate output computation from checkpoint updates: it caches recent SSM
    inputs and folds them into the checkpoint periodically, avoiding
    full-state writes on most steps. \textbf{Bottom:} three challenges of
    conventional decoding, and how ReplaySSM addresses each. (1)~SSM layers can
    bottleneck hybrid models up to 100K tokens (Nemotron-3-Super-120B-A12B-NVFP4,
    batch size 256, one B300); ReplaySSM roughly halves their state traffic. (2)~State rollback, expensive to support
    in conventional speculative decoding, becomes a buffer-pointer move. (3)~A
    low-precision checkpoint serves outputs while a high-precision one carries
    the recurrence, so rounding error no longer re-enters the recurrence.}
    \label{fig:replayssm-overview}
\end{figure}

A fixed-size state, however, does not make SSM decoding cheap. Current SSM
decoding follows the recurrence step by step. Each step loads the state,
updates it, computes the output from the updated state, and stores it back. This leads to three challenges. First, the update has
low arithmetic intensity. Each step moves the full state for a decay and a
rank-1 update. At serving batch sizes, this state traffic can make SSM updates
the bottleneck up to 100K tokens even in hybrid models whose attention cost
grows with output context length (Figure~\ref{fig:replayssm-overview}, bottom left).
Second, an updated state has no cheap rollback. This becomes a problem for
speculative decoding, which verifies multiple draft tokens together and must
restore the state to the last accepted token when a suffix is rejected
\citep{leviathan2023fast}. Serving systems such as
vLLM \citep{kwon2023efficient} therefore work around this by storing a
separate recurrent-state snapshot for each draft token. These writes slow an
already memory-bound kernel, and storing the snapshots caps concurrency, so
speculative decoding gains little at serving batch sizes. Third, quantizing
the SSM state feeds rounding error into subsequent recurrent updates, where
it can accumulate over a long generation. Aggressively
reducing state precision can therefore distort the output distribution and
degrade downstream accuracy.

These three challenges motivate the question: \emph{Why do we store an
updated state after every step?} Unrolling the recurrence shows that the
current state equals a decayed earlier checkpoint plus a short sum over
recent inputs. ReplaySSM therefore represents the state as a checkpoint and
a bounded buffer of recent SSM inputs (Figure~\ref{fig:replayssm-overview},
top). This representation separates output computation from checkpoint
updates. On most steps (non-flush steps), ReplaySSM appends the new inputs
to the buffer and computes the output from the checkpoint and the buffer
without storing a state. Only when the buffer fills (flush steps) does it
fold the buffered inputs into the checkpoint and start a new window. The
outputs match conventional decoding up to floating-point differences.

This separation addresses all three challenges. A non-flush step writes only
the new input vectors and decay factors, which are far smaller than the
state, so checkpoint writes are amortized over the window and the dominant
state traffic roughly halves. Keeping recent inputs also makes speculative
rollback cheap. Rejected draft tokens are discarded by moving a buffer
pointer, and because non-flush steps need only outputs, verification can
compute them directly from the checkpoint and the cached inputs without
materializing intermediate states. For low-precision states, the two roles
can use different precisions: a low-precision checkpoint serves output
computation on non-flush steps, while a higher-precision checkpoint carries
the recurrence across flushes, so rounding error in the low-precision copy
never enters later updates.

Our contributions are:
\nopagebreak
{\makeatletter\@beginparpenalty=10000\makeatother
\begin{itemize}
    \item We introduce ReplaySSM, which represents the SSM state as a
    checkpoint plus a buffer of recent inputs, separating output computation
    from checkpoint updates and avoiding full-state writes on most decoding
    steps (Section~\ref{sec:standard-decoding}).
    \item We extend it to speculative decoding, where the cached inputs give
    pointer-based rollback and output-only verification
    (Section~\ref{sec:replayssm-speculative}).
    \item We extend it to low-precision state storage, where the separation
    lets output computation and recurrent updates use different state
    precisions (Section~\ref{sec:replayssm-low-precision}).
    \item We implement ReplaySSM in vLLM for Mamba-2 and GDN. At serving
    batch size 256, it speeds up standard decoding by $1.20$--$1.47\times$
    over a tuned vLLM baseline across four hybrid models, and delivers about
    $1.8\times$ the throughput of vLLM's speculative decoding on the two
    120B-class mixture-of-experts (MoE) models. Separate state precisions
    also reduce output Kullback--Leibler (KL) divergence and downstream
    accuracy losses compared with a shared
    low-precision checkpoint (Section~\ref{sec:experiments}).
\end{itemize}}

\section{Preliminaries}
\label{sec:preliminaries}

\subsection{State-Space Models}
\label{sec:ssm-preliminaries}

SSMs compress the token history into a fixed-size recurrent state instead of
a growing KV cache. In Mamba-2, the recurrence is
\begin{equation}
    S_t = \alpha_t S_{t-1} + v_t k_t^\top,
    \qquad
    o_t = S_t q_t,
    \label{eq:mamba2-recurrence}
\end{equation}
where $S_t \in \mathbb{R}^{d_v \times d_k}$ is the state,
$v_t,o_t \in \mathbb{R}^{d_v}$, $k_t,q_t \in \mathbb{R}^{d_k}$, and
$\alpha_t \in (0,1)$ is a data-dependent scalar decay. We write
$\gamma_{i:j} = \prod_{\ell=i}^{j}\alpha_\ell$ for the cumulative decay, with
$\gamma_{i:j}=1$ when $i>j$, and abbreviate $\gamma_{1:h}$ as $\gamma_h$.
Each step loads $S_{t-1}$ from HBM, forms $S_t$ and $o_t$, and stores
$S_t$ back.

Delta-rule models such as GDN and KDA modify stored information before writing new information. The GDN
recurrence is
\begin{equation}
    S_t = \alpha_t S_{t-1}
    \left(I-\beta_t k_t k_t^\top\right)
    + \beta_t v_t k_t^\top,
    \label{eq:gdn-recurrence}
\end{equation}
where $\beta_t \in (0,1)$ controls the write strength. With
\begin{equation}
    u_t = \beta_t\left(v_t-\alpha_t S_{t-1}k_t\right)
    \qquad\text{it becomes}\qquad
    S_t = \alpha_t S_{t-1} + u_t k_t^\top,
    \label{eq:gdn-rewrite}
\end{equation}
the same outer-product form as Mamba-2 with $u_t$ in place of $v_t$, at the
cost of computing $u_t$ from the previous state.

\subsection{Speculative Decoding and Quantization}
\label{sec:speculative-decoding-preliminaries}
\label{sec:quantization-preliminaries}

Speculative decoding uses a lightweight drafter to propose multiple candidate
tokens, which the target model verifies in one forward pass; how many are
accepted is known only after this pass. Rejected tokens are discarded, and the KV cache or SSM state must retain only the accepted
prefix \citep{leviathan2023fast}. Quantization stores values in a lower-precision format
to cut storage and memory traffic. Values the format cannot represent are
rounded: round-to-nearest (RTN) can introduce systematic bias, while
stochastic rounding (SR) is unbiased in expectation but still has nonzero
variance \citep{gupta2015deep,croci2022stochastic}.

\section{Motivation and Key Insight}
\label{sec:motivation}

SSMs compress an $O(N)$-token history into an $O(1)$-size state through
recurrent updates. However, directly and eagerly following the recurrence,
materializing the updated state at every decoding step, creates three
challenges (Figure~\ref{fig:replayssm-overview}, bottom): low arithmetic
intensity, expensive rollback, and propagation of quantization error through
the recurrence.

\subsection{Challenge 1: Large State I/O with Little Compute}
\label{sec:challenge-state-io}

Conventional decoding reads and writes the full recurrent state at each step
while doing little computation on it; Mamba-2, for example, applies only an
elementwise decay and a rank-1 update. State updates are therefore
memory-bound, and state I/O dominates their latency
(Appendix~\ref{app:analysis:intensity}). One might expect attention to dominate hybrid-model latency because its
KV-cache traffic grows with context length. However, for
Nemotron-3-Super-120B-A12B-NVFP4 at batch size 256 on one B300, SSM state
updates remain the largest latency component up to 100K tokens
(Figure~\ref{fig:replayssm-overview}, bottom left). Two factors explain
this. The model has five times as many SSM layers as attention layers, and
each SSM layer's FP32 state is 4~MiB per sequence, the size of an attention
layer's FP8 KV cache for 8K tokens. A decoding step loads and stores the
state but only loads the KV cache, so the SSM layers move more bytes than
the attention layers until the context reaches about 80K tokens
(Appendix~\ref{app:analysis:crossover}). The measured crossover
near 100K tokens is consistent with attention kernels having received more
optimization effort \citep[e.g.,][]{dao2022flashattention}.

\subsection{Challenge 2: Recurrent State Has No Cheap Rollback}
\label{sec:challenge-rollback}

Conventional SSM decoding eagerly summarizes each token's inputs into the
state at once, and the update has no cheap inverse. This matters for
speculative decoding, where the target model must roll the state back when
it rejects a suffix of draft tokens (Figure~\ref{fig:replayssm-overview},
bottom middle). Attention handles rollback naturally by moving the KV-cache
pointer to the end of the accepted prefix. An SSM state contains no token
history to point into, so serving systems such as vLLM store a state snapshot for each draft token and
select the one matching the accepted prefix. Verifying $T$ tokens costs
$T$ full-state writes per step on an already memory-bound path and reduces
serving concurrency. Prior work (\citealp{wang2024mamba,wu2024snakes,wu2026stree};
Section~\ref{sec:related-work}) avoids the per-token snapshots but still
pays extra state traffic to recover the accepted state by re-running the
recurrence, and at a serving batch size none of it is more than a few
percent faster than vLLM's snapshots (Appendix~\ref{app:prior-methods}).
Speculative decoding therefore gives SSMs little speedup at serving batch
sizes.

\subsection{Challenge 3: Quantization Error Propagates Through Recurrence}
\label{sec:challenge-quantization}

In low-precision SSM decoding, the stored quantized state both produces the
current output and feeds the next recurrent update. Rounding error therefore re-enters
the recurrence at every step and can accumulate with sequence length
(Figure~\ref{fig:replayssm-overview}, bottom right). Round-to-nearest adds systematic bias; stochastic rounding is unbiased but
still propagates its variance (Appendix~\ref{app:analysis:error}).

\subsection{ReplaySSM: Cache SSM Inputs, Not State}
\label{sec:replayssm-insight}

All three problems come from one state-management policy: eagerly summarize each
token's inputs into the state at once and write the state to HBM at every
step. The recurrence itself shows that this is unnecessary. For Mamba-2, let
$S_{\mathrm{ckpt}}$ be the state before a window of $h$ inputs indexed
locally by $j=1,\ldots,h$. Then
\begin{equation}
    S_t = \alpha_t S_{t-1} + v_t k_t^\top
    = \gamma_h S_{\mathrm{ckpt}}
    + \sum_{j=1}^{h} \gamma_{j+1:h} v_j k_j^\top .
    \label{eq:mamba2-unrolled}
\end{equation}
The recurrent form loads the summary $S_{t-1}$ and applies the current
inputs; the unrolled form reconstructs the same state from
$S_{\mathrm{ckpt}}$ and the recent inputs $\{(\alpha_j,v_j,k_j)\}_{j=1}^{h}$.
Delta-rule models unroll the same way (Appendix~\ref{app:gdn:derivation})
once their updates are written in outer-product form
(Equation~\ref{eq:gdn-rewrite}).

ReplaySSM therefore represents the state as a checkpoint $S_{\mathrm{ckpt}}$
plus a bounded buffer $\mathcal{B}$ of recent SSM inputs; for Mamba-2 each
record holds $(\alpha_j,v_j,k_j)$. This representation separates output
computation from checkpoint updates. Conventional decoding fuses the two,
because the output is read from the state it has just updated and must store
for the next step. A non-flush step appends the current
inputs to $\mathcal{B}$ and computes the output from $S_{\mathrm{ckpt}}$ and
the buffer without storing a state. When the buffer reaches its capacity, a
flush step folds the buffered inputs into $S_{\mathrm{ckpt}}$, computes the
output, writes the new checkpoint to HBM, and clears the buffer. State writes
therefore happen only on flush steps, and the outputs equal those of
conventional decoding up to floating-point differences
(Figure~\ref{fig:replayssm-overview}). Sections~\ref{sec:standard-decoding}
to~\ref{sec:replayssm-low-precision} build on this separation for standard
decoding, speculative decoding, and low-precision state, and
Section~\ref{sec:related-work} reviews related work.

\section{ReplaySSM for Standard Decoding}
\label{sec:standard-decoding}

Standard decoding is the simplest setting in which to see how the
separation reduces state memory traffic (Challenge~1). The core operation reconstructs the current
state from a checkpoint and the cached inputs. For a checkpoint
$S_{\mathrm{ckpt}}$ and replay buffer
$\mathcal{B}_{1:h}=\{(\alpha_j,v_j,k_j)\}_{j=1}^{h}$, define
\begin{equation}
    \textsc{Replay}(S_{\mathrm{ckpt}},\mathcal{B}_{1:h})
    = \gamma_h S_{\mathrm{ckpt}}
    + \sum_{j=1}^{h}\gamma_{j+1:h}v_jk_j^\top.
    \label{eq:replay}
\end{equation}

\begin{figure}[t]
\small
\definecolor{algchg}{RGB}{0,84,170}
\providecommand{\chg}[1]{\textcolor{algchg}{#1}}
\providecommand{\algpanelhead}[2]{%
    \refstepcounter{algorithm}\label{#1}%
    \textbf{Algorithm~\thealgorithm}\enspace #2\par
    \vspace{1.5pt}\hrule\vspace{3pt}}
\providecommand{\algpanelrule}{\par\vspace{4pt}\hrule\vspace{4pt}}
\algrenewcommand{\algorithmicindent}{0.9em}
\algtext*{EndIf}
\providecommand{\algspread}{\linespread{1.18}\selectfont} 

\hrule height 0.8pt
\vspace{3pt}
\noindent
\textbf{ReplaySSM decoding step (Mamba-2).}
Algorithm~\ref*{alg:replayssm-standard} is standard decoding.
Algorithm~\ref*{alg:replayssm-speculative} extends it to verify $T$
tokens per step and commits only the accepted prefix.
Algorithm~\ref*{alg:replayssm-low-precision} adds a second checkpoint
precision, with its changes from Algorithm~\ref*{alg:replayssm-speculative}
in \chg{color}.
\algpanelrule

\noindent
\begin{minipage}[t]{0.47\linewidth}
\vspace{0pt}\raggedright
\algpanelhead{alg:replayssm-standard}{Standard decoding}
\begin{algorithmic}[1]\algspread
    \Statex \textbf{Input:} current token's $x=(\alpha,v,k)$ and $q$
    \State $\mathcal{B}_{h+1}\gets x$; $h\gets h+1$
    \State $S\gets\textsc{Replay}(S_{\mathrm{ckpt}},\mathcal{B}_{1:h})$
    \State $o\gets Sq$
    \If{$h=L$} \Comment{flush}
        \State $S_{\mathrm{ckpt}}\gets S$; store $S_{\mathrm{ckpt}}$ to HBM
        \State $\mathcal{B}\gets\varnothing$; $h\gets0$
    \EndIf
    \State \Return $o$
\end{algorithmic}
\end{minipage}\hfill
\begin{minipage}[t]{0.505\linewidth}
\vspace{0pt}
\textbf{Notation} (full definitions in Appendix~\ref{app:notation})\par
\vspace{1.5pt}\hrule\vspace{3pt}
\linespread{1.12}\selectfont
\providecommand{\notationitem}[1]{\par\hangindent=1em\hangafter=1\noindent #1\ }
\notationitem{$S_{\mathrm{ckpt}}$:} checkpoint state in HBM
    (Alg.~\ref*{alg:replayssm-low-precision}: $S_{\mathrm{ckpt}}^{\mathrm{HP}}$, $S_{\mathrm{ckpt}}^{\mathrm{LP}}$)
\notationitem{$S$:} working state on chip
\notationitem{$\mathcal{B}_{i:j}$:} replay-buffer records $i$ to $j$, each $x=(\alpha,v,k)$
\notationitem{$h$, $L$:} $\mathcal{B}_{1:h}$ is committed, $h\le L$; window size $L$
\notationitem{$X$, $Q$:} inputs $\{x_s\}_{s=1}^{T}$ and queries $[q_s]_{s=1}^{T}$ of $T$ tokens
\notationitem{\textsc{Replay}:} state from checkpoint and records (Eq.~\ref{eq:replay})
\notationitem{\textsc{OutputOnly}:} outputs for $Q$ without states (Eq.~\ref{eq:output-only})
\notationitem{$\mathcal{Q}_{\mathrm{HP}}$, $\mathcal{Q}_{\mathrm{LP}}$:} casts to the two checkpoint formats
\end{minipage}
\algpanelrule

\noindent
\begin{minipage}[t]{0.47\linewidth}
\vspace{0pt}\raggedright
\algpanelhead{alg:replayssm-speculative}{Speculative decoding}
\begin{algorithmic}[1]\algspread
    \Statex \textbf{Input:} $X$ and $Q$ of $T$ tokens
    \State $\mathcal{B}_{h+1:h+T}\gets X$ \Comment{$h$ unchanged}
    \Statex \strut
    \State Load $S_{\mathrm{ckpt}}$ from HBM into $S$
    \State $O\gets\textsc{OutputOnly}(S,\mathcal{B}_{1:h+T},Q)$
    \If{$h+T>L$} \Comment{flush committed records}
        \State $S\gets\textsc{Replay}(S,\mathcal{B}_{1:h})$
        \State $S_{\mathrm{ckpt}}\gets S$
        \State Store $S_{\mathrm{ckpt}}$ to HBM
        \State Drop $\mathcal{B}_{1:h}$; $h\gets0$ \Comment{keep $\mathcal{B}_{h+1:h+T}$}
    \EndIf
    \State \Return $O$
    \Statex Later, given accepted length $a$:
    \State $h\gets h+a$ \Comment{commit accepted drafts}
\end{algorithmic}
\end{minipage}\hfill
\begin{minipage}[t]{0.505\linewidth}
\vspace{0pt}\raggedright
\algpanelhead{alg:replayssm-low-precision}{Two-precision speculative decoding}
\begin{algorithmic}[1]\algspread
    \Statex \textbf{Input:} $X$ and $Q$ of $T$ tokens
    \State $\mathcal{B}_{h+1:h+T}\gets X$ \Comment{$h$ unchanged}
    \State \chg{$\pi\gets\mathrm{HP}$ if $h+T>L$, else $\mathrm{LP}$}
    \State \chg{Load $S_{\mathrm{ckpt}}^{\pi}$ from HBM into $S$}
    \State $O\gets\textsc{OutputOnly}(S,\mathcal{B}_{1:h+T},Q)$
    \If{$h+T>L$} \Comment{flush committed records}
        \State $S\gets\textsc{Replay}(S,\mathcal{B}_{1:h})$
        \State \chg{$S_{\mathrm{ckpt}}^{\mathrm{HP}}\gets\mathcal{Q}_{\mathrm{HP}}(S)$;
            $S_{\mathrm{ckpt}}^{\mathrm{LP}}\gets\mathcal{Q}_{\mathrm{LP}}(S)$}
        \State \chg{Store $S_{\mathrm{ckpt}}^{\mathrm{HP}}$ and
            $S_{\mathrm{ckpt}}^{\mathrm{LP}}$ to HBM}
        \State Drop $\mathcal{B}_{1:h}$; $h\gets0$ \Comment{keep $\mathcal{B}_{h+1:h+T}$}
    \EndIf
    \State \Return $O$
    \Statex Later, given accepted length $a$:
    \State $h\gets h+a$ \Comment{commit accepted drafts}
\end{algorithmic}
\end{minipage}
\par\vspace{3pt}
\hrule height 0.8pt
\end{figure}

Algorithm~\ref{alg:replayssm-standard} applies this operation once per token.
Lines~1--3 append the current inputs, reconstruct the state, and compute the
output; the only full-state store is Line~5, inside the flush branch.
Algorithms~\ref{alg:replayssm-speculative}
and~\ref{alg:replayssm-low-precision} extend it in
Sections~\ref{sec:replayssm-speculative}
and~\ref{sec:replayssm-low-precision}. For GDN, each record stores the correction
$u_j$ in place of $v_j$ and the flush replays the same sum
(Appendix~\ref{app:gdn:standard}).

Conventional decoding reads and writes the full recurrent state at every
step. ReplaySSM still reads $S_{\mathrm{ckpt}}$ at each step, since the output
computation needs it, but writes the state only when the checkpoint is
updated. A non-flush step only appends the current inputs to $\mathcal{B}$.
Over an $L$-step window, both methods perform $L$ full-state reads, while
conventional decoding performs $L$ full-state writes and ReplaySSM performs
one, plus the buffer reads and writes. Because the buffer is much smaller
than the state, ReplaySSM roughly halves the dominant state memory
traffic (Appendix~\ref{app:analysis:traffic}).

The window size $L$ balances checkpoint writes against replay overhead. A
larger $L$ amortizes each checkpoint write over more steps but increases the
number of buffered inputs loaded and applied at each step. We use $L=16$
throughout (Appendix~\ref{app:analysis:window}).

\section{ReplaySSM for Speculative Decoding}
\label{sec:replayssm-speculative}

\begin{table}[t]
    \sbox0{\includegraphics[width=2.65in]{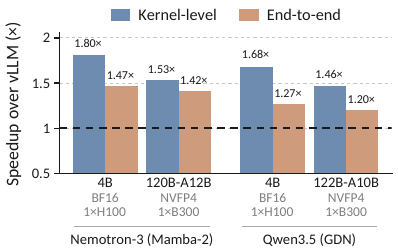}}%
    \begin{minipage}[t]{0.485\linewidth}
        \centering
        \parbox[b][\ht0][c]{\linewidth}{%
            \centering
            \footnotesize
            \setlength{\tabcolsep}{3pt}
            \renewcommand{\arraystretch}{1.5}
            \begin{tabular}{@{}lll@{}}
                \toprule
                Model & SSM & Weights \\
                \midrule
                Nemotron-3-Nano-4B & Mamba-2 & BF16 \\
                Nemotron-3-Super-120B-A12B & Mamba-2 & NVFP4 \\
                Qwen3.5-4B & GDN & BF16 \\
                Qwen3.5-122B-A10B & GDN & NVFP4 \\
                \bottomrule
            \end{tabular}}
        \tightcaption
        \caption{Evaluated models. The 4B models run on one H100 and the
        120B-class models on one B300. More details are in
        Table~\ref{tab:exp:model-geometry}.}
        \label{tab:exp:model-settings}
    \end{minipage}\hfill
    \begin{minipage}[t]{0.485\linewidth}
        \centering
        \usebox0
        \tightcaption
        \captionof{figure}{Standard-decoding speedup over tuned conventional
        vLLM ($B=256$, $L=16$, FP32 state). Kernel numbers include flush
        costs.}
        \label{fig:exp:standard-decoding-speedup}
    \end{minipage}
\end{table}

In speculative decoding, ReplaySSM keeps the traffic saving of standard
decoding (Section~\ref{sec:standard-decoding}) and changes how the target
model verifies draft tokens in two further ways:
cached inputs make
rollback cheap, and output-only evaluation avoids materializing intermediate
recurrent states.

\subsection{Cheap Rollback with Cached Inputs}
\label{sec:speculative-rollback}

The accepted-prefix length is known only after target-model verification
(Section~\ref{sec:speculative-decoding-preliminaries}). ReplaySSM therefore
appends the SSM inputs of all $T$ tokens of the verification pass to the replay buffer first and
commits only the accepted prefix after verification by moving the buffer
pointer. Algorithm~\ref{alg:replayssm-speculative} gives the procedure for
Mamba-2. The draft records are written after the $h$ committed records
(Line~1), and all $h+T$ records feed the output computation (Line~3). The
accepted length $a$ is known only after the outputs are returned (Line~9),
and it then advances $h$ (Line~10). Rejected records lie beyond position $h$
and are overwritten by the next append. The flush branch (Lines~4--8) folds
only committed inputs into the checkpoint, so it can run before acceptance
is known. Non-flush steps write only the small input records, and only flush
steps write the checkpoint. Appendix~\ref{app:prior-methods} compares the
per-step state traffic with vLLM's per-draft snapshots and with the prior
recovery methods of Section~\ref{sec:challenge-rollback}.

We implement $\mathcal{B}$ as a circular buffer with capacity $L+T$, room for
$L$ committed records and $T$ drafts, so dropping the committed prefix
(Line~8) advances the ring origin instead of copying the draft records
forward. Flushing when
$h+T>L$ keeps the committed length at most $L$ after verification
(Appendix~\ref{app:implementation:flush}). Unlike the fixed $L$-step
interval of standard decoding, the flush frequency now depends on
acceptance, since higher acceptance fills the buffer faster.

\subsection{Output-Only Verification}
\label{sec:output-only-verification}

The separation also lets a non-flush step avoid materializing the state
as well as storing it. Conventional decoding must produce the updated state
because the next step consumes it. Under ReplaySSM no later step reads the state of a non-flush
step, so that step needs only the output. For a single Mamba-2 update from a zero state, the product
of $v$, $k^\top$, and $q$ can be bracketed two ways,
$o=(vk^\top)q=v(k^\top q)$: the left side materializes the $d_v\times d_k$
state, while the right side forms the scalar $k^\top q$ and scales $v$. We
call the right-hand form output-only evaluation.

For verification, let $V$ and $K$ hold the $h+T$ buffered values and keys as
columns, committed records first, and let $Q=[q_1,\ldots,q_T]$. The same idea
gives
\begin{equation}
    \textsc{OutputOnly}(S_{\mathrm{ckpt}},\mathcal{B}_{1:h+T},Q)
    = S_{\mathrm{ckpt}}QD + V\bigl(W\odot(K^\top Q)\bigr),
    \label{eq:output-only}
\end{equation}
where $D$ is diagonal with $D_{ss}=\gamma_{h+s}$, $\odot$ is the elementwise
product, and $W$ holds cumulative decays under a causal mask,
$W_{js}=\gamma_{j+1:h+s}$ for $j\le h+s$ and $0$ otherwise. The mask
restricts each draft query to the committed history and the drafts up to its
own position. The output-only form has two benefits. It never materializes a state, so
it needs fewer FLOPs (Appendix~\ref{app:analysis:flops}), and $K^\top Q$ is
computed once for every group of value heads that share keys and queries.
Both matter more in speculative decoding, where verifying $T$ tokens
multiplies the arithmetic per step. The outputs match recurrent evaluation
up to floating-point differences. GDN admits the same form. Its draft
corrections depend on the earlier drafts in the window, and
Appendix~\ref{app:gdn:speculative} obtains all $T$ of them from one
$T\times T$ triangular solve. Equation~\ref{eq:output-only} also has the
chunkwise form of prefill and training kernels, with the buffered inputs
followed by the drafts as the chunk (Section~\ref{sec:related-work}).

\section{ReplaySSM for Low-Precision State}
\label{sec:replayssm-low-precision}

\begin{figure}[t]
    \sbox0{\includegraphics[width=2.65in]{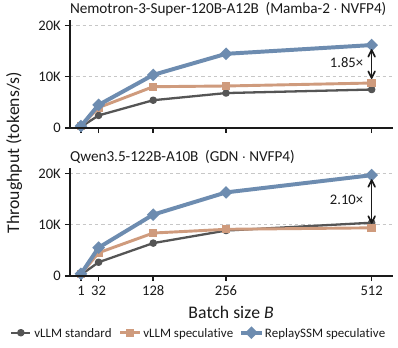}}%
    \begin{minipage}[t]{0.485\linewidth}
        \centering
        \usebox0
        \tightcaption
        \caption{Speculative serving throughput ($L=16$, $T=4$, FP32 state,
        one B300 per model).}
        \label{fig:exp:speculative-serving-throughput}
    \end{minipage}\hfill
    \begin{minipage}[t]{0.485\linewidth}
        \centering
        \includegraphics[height=\ht0]{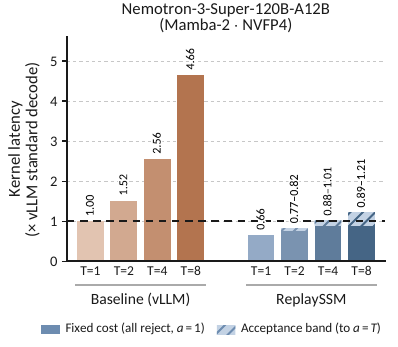}
        \tightcaption
        \caption{Speculative kernel latency ($B=256$, $L=16$, FP32 state),
        flushes included.}
        \label{fig:exp:speculative-kernel-latency-mamba2}
    \end{minipage}
\end{figure}

Lower-precision state storage reduces memory traffic and can reduce latency
in memory-bound SSM kernels. In conventional decoding the same stored state computes the output and
feeds the next update, so its rounding error re-enters the recurrence at
every step (Section~\ref{sec:challenge-quantization}). The separation
changes this in two ways. It quantizes the state once per window instead of once per step, though
with a single low-precision checkpoint that rounding error still enters later
updates. More important, non-flush steps read the state only to compute
outputs, while flush steps update the checkpoint that carries the recurrence
forward, so the two roles can use different precisions.

We therefore keep a high-precision
checkpoint $S_{\mathrm{ckpt}}^{\mathrm{HP}}$ for recurrent updates and a
low-precision copy $S_{\mathrm{ckpt}}^{\mathrm{LP}}$ for non-flush output
computation. Both represent the same checkpoint history;
$\mathcal{Q}_{\mathrm{HP}}$ and $\mathcal{Q}_{\mathrm{LP}}$ denote the casts
from accumulation precision. Algorithm~\ref{alg:replayssm-low-precision}
extends Algorithm~\ref{alg:replayssm-speculative} by
choosing which checkpoint to load (Lines~2--3). Non-flush steps read the
low-precision copy. Flush steps read the high-precision checkpoint,
reconstruct the committed state, and cast it independently into both formats
(Lines~7--8). The extra low-precision write on flush steps lets every
non-flush step read the smaller checkpoint. Acceptance, rollback, and buffer
management are unchanged.

Separate precisions break the feedback loop of
Section~\ref{sec:challenge-quantization}. Each flush starts from the
high-precision checkpoint and writes a fresh high-precision checkpoint; the
low-precision copy is derived from it and never read back into the
recurrence, so its rounding error does not accumulate. Rounding error in the
high-precision path remains, and non-flush outputs still depend on the
approximate copy. Appendix~\ref{app:analysis:error} provides the analysis, and
Section~\ref{sec:exp-low-precision} measures the resulting output error as
KL divergence. GDN needs one adjustment. The corrections $u$ cached on
non-flush steps come from the low-precision checkpoint
(Equation~\ref{eq:gdn-rewrite}), so we also cache $v$ and recompute the
corrections from the high-precision state at flush time, keeping
low-precision error out of the recurrence
(Appendix~\ref{app:gdn:precision}).

\section{Evaluation}
\label{sec:evaluation}
\label{sec:experiments}

\subsection{Setup}
\label{sec:experimental-setup}

We evaluate four hybrid models from two families, Nemotron-3 with Mamba-2
layers and Qwen3.5 with GDN layers (Table~\ref{tab:exp:model-settings}),
each on one GPU without tensor parallelism. We implement ReplaySSM in vLLM
(Appendix~\ref{app:implementation}) and compare against tuned conventional
vLLM decoding with CUDA graphs. All performance experiments store the
recurrent state in FP32, the default for all four models.
Section~\ref{sec:exp-low-precision} lowers only the
state precision and leaves weights as released (Table~\ref{tab:exp:model-settings}). We write $B$ for batch size, $L$ for the replay window
size, $T$ for the verification length, the tokens verified per step ($T=1$ is
standard decoding), $a$
for the number of those tokens it commits, $1\le a\le T$, and $N$ for
generation length. Standard decoding reports
whole-model latency per decode step and standalone SSM-kernel latency.
Speculative decoding reports serving throughput in generated tokens per
second, prefill included, and standalone kernel latency for verification.
The low-precision experiments measure the KL divergence
$D_{\mathrm{KL}}(p_{\mathrm{FP32}} \Vert p_{\mathrm{config}})$ from
conventional FP32-state decoding under teacher forcing on PG-19
\citep{rae2019compressive} for $N$ up to 128K tokens, and generated-answer accuracy on MMLU-Pro \citep{wang2024mmlu},
GPQA-Diamond \citep{rein2023gpqa}, and the Level~5 subset of MATH
\citep{hendrycks2021measuring}, Math-L5. Details are in
Appendix~\ref{app:exp:setup}.

\begin{figure}[t]
    \begin{minipage}[t]{0.545\linewidth}
        \centering
        \includegraphics[width=2.95in]{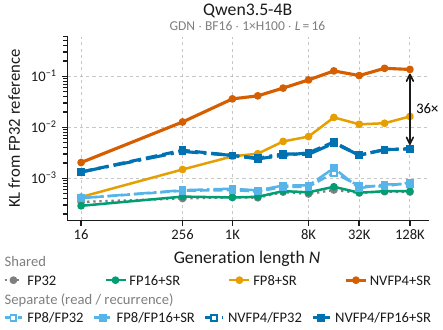}
        \tightcaption\vspace{3pt}
        \caption{Output KL from conventional FP32-state decoding under
        teacher forcing on PG-19.}
        \label{fig:exp:precision-kl-small-models}
    \end{minipage}\hfill
    \begin{minipage}[t]{0.435\linewidth}
        \centering
        \includegraphics[width=2.35in]{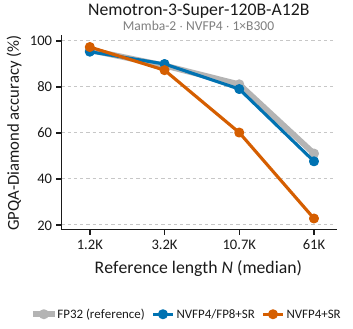}
        \tightcaption\vspace{3pt}
        \caption{GPQA-Diamond accuracy by FP32 reference-length quartile.}
        \label{fig:exp:precision-gpqa-by-length}
    \end{minipage}
\end{figure}

\subsection{Standard Decoding}
\label{sec:exp-standard-decoding}

The baseline uses the fastest kernel launch configuration for each model,
which beats the shipped Mamba-2 defaults (Appendix~\ref{app:exp:batch}).
At $B=256$ and $L=16$, ReplaySSM speeds up the SSM kernel by
$1.46$--$1.80\times$ and the whole decode step by $1.20$--$1.47\times$
across the four models (Figure~\ref{fig:exp:standard-decoding-speedup}).
The kernel gain is the traffic reduction of
Section~\ref{sec:standard-decoding} and appears where the conventional
kernel is memory-bound. Appendix~\ref{app:exp:batch} sweeps the batch size
from 1 to 512 and reports the HBM bandwidth the conventional kernel reaches.
At $B=1$, the kernel is latency-bound, and ReplaySSM ends within a few
percent of the baseline end to end. From there the speedup grows with batch
size until the conventional kernel saturates the bandwidth. The end-to-end
gain is bounded by the SSM share of the decode step, since attention and the
rest of the model are unchanged, and it also grows with batch size because
that share grows. We use $L=16$ because it gives the lowest amortized kernel
latency for all four models at $B=256$ in the window sweep of
Appendix~\ref{app:exp:window}, which the traffic analysis of
Appendix~\ref{app:analysis:window} also predicts.

\subsection{Speculative Decoding}
\label{sec:exp-speculative-decoding}

We evaluate the two 120B-class models, which run with NVFP4 weights
(Table~\ref{tab:exp:model-settings}), at $L=16$ and $T=4$, drafting with
each model's MTP head on GSM8K prompts \citep{gsm8k} with 256 generated
tokens per request. Figure~\ref{fig:exp:speculative-serving-throughput}
compares serving throughput with conventional vLLM standard and speculative
decoding. Storing a state snapshot per draft token adds $O(T)$ state writes
per step and caps the concurrency that GPU memory allows, so conventional
speculation stops gaining beyond $B=128$ and on Qwen3.5-122B-A10B at $B=512$
falls below standard decoding. ReplaySSM keeps improving with batch size.
At $B=512$ it delivers $1.85\times$ and $2.10\times$ the throughput of
conventional speculation, and $2.16\times$ and $1.90\times$ that of
standard decoding, with matched acceptance lengths.
Appendix~\ref{app:exp:speculative} gives the serving setup, the throughput
over the full grid of $B$ and $T$, and the acceptance lengths. The faster
verification kernel is the other source of the gain.
Figure~\ref{fig:exp:speculative-kernel-latency-mamba2} shows its cost on
Nemotron-3-Super-120B-A12B at $B=256$. Conventional verification stores one
snapshot per token, so its latency grows with $T$ and reaches $4.66\times$ a
standard-decoding step at $T=8$. ReplaySSM verifies eight tokens for
$0.89$--$1.21\times$ of a standard step, flushes included. The band spans
all drafts rejected to all accepted, since higher acceptance fills the
buffer and flushes more often. The GDN kernel shows the same trend
(Appendix~\ref{app:exp:spec-kernel}).

\subsection{Low-Precision State}
\label{sec:exp-low-precision}
\begin{table}[t]
    \caption{Downstream accuracy. Bold: mean change from FP32 over the four
    models, in percentage points (FP32 row: accuracy). Gray: per-model values, Nemotron-3-Nano-4B /
    Nemotron-3-Super-120B-A12B-NVFP4 / Qwen3.5-4B / Qwen3.5-122B-A10B-NVFP4. Red:
    losses $\ge 5$ points.}
    \label{tab:exp:precision-accuracy}
    \centering
    \footnotesize
    \setlength{\tabcolsep}{3pt}
\providecommand{\permodel}[1]{{\scriptsize\textcolor{black!70}{(#1)}}}
\providecolor{tabloss}{HTML}{D32F2F}
\providecommand{\loss}[1]{\textcolor{tabloss}{#1}}
\begin{tabular}{@{}lccc@{}}
\toprule
Checkpoint configuration & MMLU-Pro & GPQA-Diamond & Math-L5 \\
\midrule
Conventional FP32 (reference) & $\mathbf{77.6\%}$ & $\mathbf{73.7\%}$ & $\mathbf{92.3\%}$ \\
 & \permodel{$67.2$\,/\,$81.4$\,/\,$79.0$\,/\,$82.9$} & \permodel{$56.7$\,/\,$79.3$\,/\,$76.3$\,/\,$82.7$} & \permodel{$89.5$\,/\,$95.5$\,/\,$93.7$\,/\,$90.7$} \\
\midrule
Shared FP16+SR & $\mathbf{-0.09}$ & $\mathbf{-2.02}$ & $\mathbf{0.00}$ \\
 & \permodel{$-0.3$\,/\,$+0.3$\,/\,$-0.5$\,/\,$+0.1$} & \permodel{$-2.2$\,/\,$-3.4$\,/\,$-1.5$\,/\,$-1.0$} & \permodel{$-1.1$\,/\,$+0.2$\,/\,$-0.3$\,/\,$+1.2$} \\
\addlinespace[2pt]
Shared FP8+SR & $\mathbf{-0.24}$ & $\mathbf{-1.05}$ & $\mathbf{+0.01}$ \\
 & \permodel{$-0.6$\,/\,$+0.2$\,/\,$-0.4$\,/\,$-0.1$} & \permodel{$-1.0$\,/\,$+0.0$\,/\,$-0.5$\,/\,$-2.7$} & \permodel{$-1.0$\,/\,$+0.7$\,/\,$-0.5$\,/\,$+0.9$} \\
\addlinespace[2pt]
Shared NVFP4+SR & \loss{$\mathbf{-5.88}$} & \loss{$\mathbf{-13.05}$} & \loss{$\mathbf{-10.92}$} \\
 & \permodel{\loss{$-21.0$}\,/\,$-0.9$\,/\,$-1.1$\,/\,$-0.5$} & \permodel{\loss{$-31.3$}\,/\,\loss{$-12.5$}\,/\,$-3.2$\,/\,\loss{$-5.2$}} & \permodel{\loss{$-41.7$}\,/\,$-2.9$\,/\,$-1.7$\,/\,$+2.5$} \\
\midrule
Separate FP8 / FP16+SR & $\mathbf{-0.04}$ & $\mathbf{-0.25}$ & $\mathbf{-0.12}$ \\
 & \permodel{$-0.6$\,/\,$+0.2$\,/\,$-0.4$\,/\,$+0.6$} & \permodel{$-3.5$\,/\,$+0.2$\,/\,$+2.4$\,/\,$+0.0$} & \permodel{$-0.6$\,/\,$+0.3$\,/\,$-0.3$\,/\,$+0.1$} \\
\addlinespace[2pt]
Separate NVFP4 / FP16+SR & $\mathbf{-0.15}$ & $\mathbf{-2.36}$ & $\mathbf{-0.10}$ \\
 & \permodel{$-1.7$\,/\,$+0.5$\,/\,$-0.2$\,/\,$+0.8$} & \permodel{\loss{$-6.7$}\,/\,$-1.4$\,/\,$+0.7$\,/\,$-2.0$} & \permodel{$-1.5$\,/\,$-0.1$\,/\,$-0.1$\,/\,$+1.3$} \\
\addlinespace[2pt]
Separate NVFP4 / FP8+SR & $\mathbf{-0.40}$ & $\mathbf{-1.77}$ & $\mathbf{-0.38}$ \\
 & \permodel{$-2.1$\,/\,$+0.2$\,/\,$-0.6$\,/\,$+0.8$} & \permodel{\loss{$-5.6$}\,/\,$-1.4$\,/\,$+1.5$\,/\,$-1.7$} & \permodel{$-2.8$\,/\,$+0.1$\,/\,$-0.3$\,/\,$+1.5$} \\
\bottomrule
\end{tabular}

\end{table}

A low-precision state reduces latency
(Appendix~\ref{app:exp:precision-latency}) but can distort the output
distribution and degrade downstream accuracy. We compare a shared checkpoint
(FP32, FP16, FP8, or NVFP4) that serves both output computation and
recurrent updates against separate read and recurrence checkpoints (FP8 or
NVFP4 reads with FP32 or FP16 recurrence under stochastic rounding, SR), all
at $L=16$ and with the formats of Appendix~\ref{app:exp:formats}.
Quantization applies to decoding only, and prefill computes the state at
full precision. These runs use standard decoding. Speculative decoding
changes how many tokens a step verifies, and acceptance sets when flushes
occur. Neither changes which values are rounded, so the error mechanism is
the same.
Figure~\ref{fig:exp:precision-kl-small-models} shows the
output KL from FP32-state decoding on Qwen3.5-4B. Shared FP8 and NVFP4
checkpoints accumulate error with generation length, as
Section~\ref{sec:challenge-quantization} describes, and keeping the
recurrence in FP32 removes most of it. At 128K tokens, NVFP4 reads with FP32
recurrence reduce KL from 0.136 to 0.0038, about $36\times$. What remains is
the read error itself, which does not grow with $N$ but keeps NVFP4 reads
several times above the FP32 floor. FP16+SR recurrence tracks FP32
recurrence closely, so a half-precision recurrence checkpoint suffices, and
we use it in the accuracy evaluation. Appendix~\ref{app:exp:kl} gives the
KL protocol and the curves of the other three models, which show the same
separation by model-dependent amounts. Appendix~\ref{app:exp:kl-window}
varies $L$. A shared checkpoint is requantized once per window, so its
error falls as $L$ grows (Section~\ref{sec:replayssm-low-precision}),
while the separate configurations are nearly unchanged.

Table~\ref{tab:exp:precision-accuracy} shows the downstream consequence,
with separate rows listing read / recurrence precision. A shared NVFP4+SR
checkpoint loses 13.05 points of GPQA-Diamond accuracy on average over the
four models, and 5.88 and 10.92 points on MMLU-Pro and Math-L5. NVFP4 reads
with FP16+SR recurrence cut these losses to 2.36, 0.15, and 0.10 points, and
with FP8+SR recurrence to 1.77, 0.40, and 0.38. Shared FP16 and FP8
checkpoints stay within the benchmark noise of FP32 even though their KL
sits above the floor, so accuracy resolves only the larger KL gaps.
Appendix~\ref{app:exp:accuracy} gives the per-model table with output
lengths. The loss concentrates in long generations, consistent with error
that accumulates over a generation. GPQA-Diamond has the longest generations
of the three benchmarks, and on Nemotron-3-Super-120B-A12B its top
reference-length quartile has a median of 61K tokens, against about 8K on
MMLU-Pro and Math-L5. There, the gap between shared NVFP4+SR and FP32 grows
from 0--2 points in the two shorter quartiles to 20.9 and 28.2 points in the
two longer ones, while NVFP4 reads with FP8+SR recurrence keep the latter
gaps at 2.0 and 3.4 points (Figure~\ref{fig:exp:precision-gpqa-by-length}).
Appendix~\ref{app:exp:accuracy-length} repeats this for all models and
benchmarks.

\FloatBarrier
\section{Related Work}
\label{sec:related-work}

\paragraph{Chunkwise SSM algorithms and prefix caching.}
Mamba \citep{gu2023mamba} makes the SSM recurrence input dependent and keeps
linear complexity with a hardware-aware parallel scan. Mamba-2
\citep{dao2024transformers} restricts the decay to a scalar per head and,
through structured state-space duality, organizes chunkwise computation
around matrix multiplications. GDN \citep{yang2025gated} combines that decay
with the targeted updates of the delta rule \citep{schlag2021linear}, and
KDA \citep{team2025kimi} makes the decay channelwise. These recurrences keep
chunkwise parallelism through the WY representation \citep{bischof1987wy}
and the UT transform \citep{joffrain2006accumulating}, following the
DeltaNet algorithm of \citet{yang2024parallelizing}. All of this work
optimizes compute throughput in training and prefill. ReplaySSM targets
memory-bound autoregressive decoding instead and separates output
computation from checkpoint updates, so that the state is written only at
flushes. Because non-flush steps compute outputs directly from the checkpoint
and the cached recent inputs, they can reuse these chunkwise algorithms, with
the buffered inputs followed by the current token, or by the tokens to
verify, as the chunk. Marconi \citep{pan2025marconi} adds prefix caching
for hybrid models, with admission and eviction policies that weigh the
reuse and storage cost of recurrent states. Prefix caching avoids repeated
prefill computation and ReplaySSM reduces state traffic during decoding, so
the two address different phases and can be used together.

\paragraph{SSM speculative decoding.}
Attention handles rollback by moving the sequence pointer of the KV cache
backward, and the rejected keys and values are never read again. An SSM
cannot, since once the draft tokens have been eagerly summarized into the
state, recovering the previous state is expensive. Serving systems such as
vLLM \citep{kwon2023efficient} therefore work around this by storing a
snapshot of the state after every verified token and selecting the one that
matches the accepted prefix.
This costs one read and $T$ writes per step against the one read and one
write of standard decoding, and the $T$ snapshots per request cap the
concurrency that GPU memory allows. Prior work avoids the per-token
snapshots but still pays extra state traffic to recover the accepted state,
and at a serving batch size none of it is more than a few percent faster
than the snapshots (Appendix~\ref{app:prior-methods}). The Mamba in the
Llama \citep{wang2024mamba} fuses recovery into the verification kernel,
which replays the committed tokens from cached inputs onto a checkpoint and
stores both the checkpoint and the end state, one read and two writes per
step. Snakes and Ladders \citep{wu2024snakes} re-runs the state update on
the accepted prefix in one of two ways. Activation Replay does so in a
separate launch from cached inputs. Joint Attainment appends a padded
backtracking sequence to the next verification, whose variable length also
requires one captured CUDA graph per possible acceptance outcome. Both
schemes copy between a checkpoint and a working state, three reads and
three or four writes per step. STree \citep{wu2026stree} extends the replay
to tree-structured drafts. Each step first replays the accepted tokens in
place on the checkpoint, one read and one write, then verifies the drafts
with a masked tree scan that reads that state and writes nothing, two reads
and one write per step. This overhead
means that speculative decoding, a reliable speedup for Transformers,
barely helps SSMs at serving batch sizes
(Figure~\ref{fig:exp:speculative-serving-throughput}), creating a major disadvantage.

ReplaySSM also caches inputs rather than states, but it does not use the
cache to recover a state after verification. The checkpoint plus the buffer
of recent inputs is the state representation itself, in standard as well as
speculative decoding, and it separates output computation from checkpoint
updates. The checkpoint together with any prefix of the buffer is the state
after that many accepted tokens, so recovery never re-runs a recurrence.
Rollback moves a buffer pointer, verification reads the checkpoint once and
computes the outputs from the buffer without materializing a state, and the
checkpoint is written only at a flush, once per window of $L$ committed
tokens. Most steps therefore read the state once and write nothing.
Appendix~\ref{app:prior-methods} compares the state traffic and kernel
latency of all methods on one layer shape.

\paragraph{SSM state quantization.}
Q-Mamba \citep{tianqi2025q} quantizes Mamba weights, activations, and the
recurrent state after training. For the state it proposes Decoupled Scale
Quantization, separate scales along the state and channel dimensions that
absorb outliers in the state, and Efficient Selectivity Reconstruction,
which fine-tunes about two percent of the parameters on calibration data.
Its error analysis includes the term that carries the previous step's
rounding error through the decay, but its training simulation drops that
term as small, and at inference the quantized state still feeds the next
update. Its evaluation stops at 2048 tokens. The Nemotron 3 Ultra report
\citep{blakeman2026nemotron}, to which authors of this paper contributed,
quantizes the Mamba state cache to FP16 or 8-bit with stochastic rounding
and explores
periodic cache checkpointing, which stores and quantizes the state only
every $CC$ steps and replays cached input activations to advance it in
between. This is the first benefit of ReplaySSM
(Section~\ref{sec:replayssm-low-precision}), fewer quantization events per
token, and the report finds the same round-to-nearest bias that
Appendix~\ref{app:exp:rounding} measures. It keeps one quantized state
that both produces the outputs and feeds the next update, so rounding
error still enters the recurrence, only less often. ReplaySSM removes the
term from the recurrence. The separation of output computation from the
checkpoint update lets the two roles use different precisions, so the
high-precision checkpoint carries the recurrence while the low-precision
copy is read only for outputs. What remains is the
rounding of the read copy in each output, which
Section~\ref{sec:exp-low-precision} measures out to 128K tokens. The
decoupled scales of Q-Mamba could further reduce that remaining read
error, and they apply to either checkpoint.

\section{Conclusion and Future Work}
\label{sec:conclusion}

Conventional SSM decoding reads and writes the state at every step, fusing
output computation with the state update. ReplaySSM unrolls the recurrence
to represent the state as a checkpoint plus a buffer of recent inputs, which
separates output computation from checkpoint updates. Most steps compute
outputs and write no state, and the checkpoint is written once per window.
This separation roughly halves the dominant state traffic, turns speculative
rollback into a buffer-pointer move with output-only verification, and keeps
a low-precision read checkpoint out of the recurrence, so its rounding error
does not accumulate with generation length. In vLLM at serving batch sizes,
where the conventional kernel is memory-bound, ReplaySSM speeds up standard
decoding by $1.20$--$1.47\times$ and gives about $1.8\times$ the throughput
of vLLM's speculative decoding on two 120B-class models, and separate
precisions reduce the output KL of an NVFP4 checkpoint by $13$--$36\times$
at the longest measured generation length. ReplaySSM is adopted in
large-scale production models and integrated into LLM serving systems
(Appendix~\ref{app:adoption}).

Every step still reads the checkpoint, and that read is the next target.
The checkpoint enters each step scaled by a per-head cumulative decay, so a
head whose decay has become negligible can skip the read on non-flush steps
and turn its flush into a pure write. Appendix~\ref{app:future-decay-skip}
outlines this direction.

\bibliography{iclr2027_conference}
\bibliographystyle{iclr2027_conference}

\clearpage
\appendix
\addtocontents{toc}{\protect\setcounter{tocdepth}{2}}
\renewcommand{\contentsname}{Appendix Contents}
{\small\hypersetup{linkcolor=black}\tableofcontents}
\clearpage
\renewcommand{\floatpagefraction}{.85}
\section{Speculative decoding compared with prior methods}
\label{app:prior-methods}

This appendix compares the kernel-level cost of ReplaySSM verification with
four prior ways of handling the SSM state during verification. These are
vLLM's per-token snapshots, The Mamba in the Llama \citep{wang2024mamba},
the two variants of Snakes and Ladders \citep{wu2024snakes}, and STree
\citep{wu2026stree}. Every method runs on one B300 at the Mamba-2 layer
shape of Nemotron-3-Super-120B-A12B with an FP32 state, batched over $B=256$
requests with one draft chain per request. One state, one request in one
layer, is $4.2$\,MB, so the batch holds $1.07$\,GB of state per layer. $T$ counts the tokens processed per verification step
and $a$ the committed tokens. Tree drafting is outside this comparison, so
STree runs with a single chain and its tree mask reduces to a causal mask.

The prior kernels come from the authors' repositories, STree at commit
\texttt{b9a0c24}\footnote{\url{https://github.com/wyc1997/stree}} and The
Mamba in the Llama at commit
\texttt{b702a3e}\footnote{\url{https://github.com/itsdaniele/speculative_mamba}},
with only the changes needed to run them at this shape under the current
Triton. Snakes and Ladders has no repository of its own. The STree
repository, which shares its first author, implements both variants, and we
use that implementation. For each method we time the kernel launches its
implementation issues in one verification step.

All baselines are tuned. Both prior repositories hard-code one launch
configuration and vLLM's kernel picks its own from a heuristic, so we sweep
the launch configuration of every kernel and report each at its
best.\footnote{At $T=4$ and $B=256$, tuning lowers the per-step kernel
latency of STree from $1856$ to $1085$\,$\mu$s, The Mamba in the Llama from
$2261$ to $923$, Snakes and Ladders from $2955$ to $1417$ (Activation
Replay) and from $4898$ to $2304$ (Joint Attainment), and vLLM's kernel from
$1043$ to $831$. Across $T=2$ to $8$ the gain is $1.4$--$3.0\times$ for the
prior kernels and $1.2$--$1.4\times$ for vLLM's. STree's released path takes
$2079$\,$\mu$s at $T=4$ against $1085$ for its algorithm.} STree is also
measured as its algorithm, the masked replay plus a read-only tree scan,
rather than as its released path, which adds two state copies that are
identities for one chain and would make it $1.7$--$2.1\times$ slower. It
runs with a chunk size of 16 instead of its default of 256, which would pad
the $T$-token window to 256 positions.

\subsection{Per-step operations and state traffic}
\label{app:prior-methods:traffic}

R and W denote one full read and one full write of the state.
\begin{itemize}
    \item \textbf{vLLM.} One launch. The kernel loads the state once and
        stores the state after each of the $T$ tokens to that token's slot,
        so the accepted token's snapshot is the next state and no recovery
        pass is needed. $1\mathrm{R}+T\mathrm{W}$, independent of
        acceptance.
    \item \textbf{The Mamba in the Llama.} One fused launch. It loads the
        checkpoint, or the end state when every draft was accepted, replays
        the tokens committed by the previous step from cached inputs, stores
        the new checkpoint, runs the $T$ current tokens, and stores the end
        state. $1\mathrm{R}+2\mathrm{W}$ at every acceptance, since the
        checkpoint store is issued even when the replay is skipped, and a
        serial chain of up to $2T-1$ steps.
    \item \textbf{Snakes and Ladders, Activation Replay.} Three launches.
        A multi-step scan re-runs the committed tokens in place on the
        checkpoint, the checkpoint is copied to a working buffer, and the
        same kernel verifies in place on that buffer. Each use of the kernel
        loads and stores the state, so $3\mathrm{R}+3\mathrm{W}$. The
        masked replay always runs $T$ steps, so the cost does not depend on
        acceptance. This path is the STree paper's ``Vanilla SD'' baseline.
    \item \textbf{Snakes and Ladders, Joint Attainment.} The same kernel
        without an activation cache. Each request contributes two rows to
        one launch, a padded backtracking row that re-processes the
        committed tokens and a draft row that continues into the drafts,
        both starting from the state verified two steps earlier. A gather
        then copies the backtracking row's state into both rows.
        $3\mathrm{R}+4\mathrm{W}$, and $2T-1$ serial steps once the batch
        is padded to one length.
    \item \textbf{STree.} Two launches. The masked replay runs in place on
        the checkpoint, then a chunk-scan kernel verifies the drafts, reading
        the state as its initial state and storing none.
        $2\mathrm{R}+1\mathrm{W}$, independent of acceptance. The released
        path adds a beam-select self-copy and a checkpoint-to-working copy,
        $5\mathrm{R}+4\mathrm{W}$ in total.
    \item \textbf{ReplaySSM.} A non-flush step reads the checkpoint once and
        computes the outputs from it and the input buffer ($1\mathrm{R}$). A
        flush step, once per about $L/a$ steps, also writes the new
        checkpoint ($1\mathrm{R}+1\mathrm{W}$).
\end{itemize}
Table~\ref{tab:app:state-traffic} sets these counts against the bytes Nsight
measures for one step.

\begin{table}[t]
    \caption{SSM-state traffic per layer per verification step at $B=256$,
    where one state is $4.2$\,MB and the batch holds $1.07$\,GB. Model:
    reads and writes times $1.07$\,GB. Nsight: DRAM bytes at base clock with
    a cold L2, summed over the launches of one step, for $T=2$ to $8$. The
    two state copies of the released STree path are memory copies rather
    than kernels, so their bytes are taken from the model.}
    \label{tab:app:state-traffic}
    \centering
    \footnotesize
    \setlength{\tabcolsep}{4pt}
    \begin{tabular}{@{}lccc>{\raggedright\arraybackslash}p{3.1cm}@{}}
        \toprule
        Method & R/W & Model (GB) & Nsight (GB) & Note \\
        \midrule
        vLLM snapshots & $1\mathrm{R}+T\mathrm{W}$ & $3.2$--$9.7$ & $3.22$--$9.73$ & grows with $T$ \\
        The Mamba in the Llama & $1\mathrm{R}+2\mathrm{W}$ & $3.2$ & $3.23$--$3.45$ & one fused launch \\
        Snakes and Ladders, Activation Replay & $3\mathrm{R}+3\mathrm{W}$ & $6.4$ & $6.37$--$6.55$ & replay, copy, verify \\
        Snakes and Ladders, Joint Attainment & $3\mathrm{R}+4\mathrm{W}$ & $7.5$ & $7.46$--$7.70$ & two rows, then a gather \\
        STree, algorithm & $2\mathrm{R}+1\mathrm{W}$ & $3.2$ & $3.22$--$3.35$ & replay, read-only verify \\
        STree, released path & $5\mathrm{R}+4\mathrm{W}$ & $9.7$ & $9.67$--$9.79$ & two extra copies \\
        ReplaySSM, non-flush & $1\mathrm{R}$ & $1.1$ & $1.18$--$1.28$ & no state store \\
        ReplaySSM, flush & $1\mathrm{R}+1\mathrm{W}$ & $2.1$ & $2.22$--$2.30$ & once per $\sim L/a$ steps \\
        \bottomrule
    \end{tabular}
\end{table}

\subsection{Results}
\label{app:prior-methods:results}

\begin{figure}[t]
    \centering
    \includegraphics[width=\linewidth]{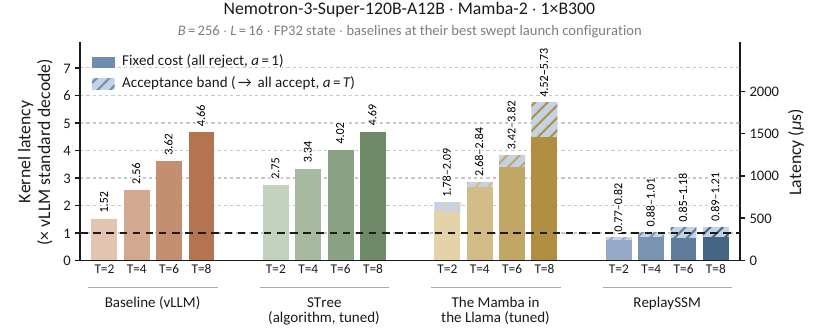}
    \caption{Per-step speculative kernel latency against prior methods on
    Nemotron-3-Super-120B-A12B (Mamba-2; $B=256$, $L=16$, FP32 state, one
    B300), normalized to conventional standard decoding ($T=1$). STree is
    measured as its algorithm, replay plus read-only tree scan without the
    released path's two state copies, and every kernel runs at its best
    swept launch configuration. STree does not depend on acceptance. Each
    band starts at the all-rejected cost ($a=1$) and ends at the most
    expensive outcome, $a=T$ for ReplaySSM, whose bars include flushes, and
    $a=T-1$ for The Mamba in the Llama, whose replay is skipped when every
    draft is accepted.}
    \label{fig:exp:speculative-kernel-latency-prior-methods}
\end{figure}

Figure~\ref{fig:exp:speculative-kernel-latency-prior-methods} adds STree
and The Mamba in the Llama to
Figure~\ref{fig:exp:speculative-kernel-latency-mamba2}. Neither beats
vLLM's snapshots at this batch size. STree costs $1.81\times$ the vLLM
kernel at $T=2$ and $1.00\times$ at $T=8$, and The Mamba in the Llama is
below it by about $6\%$ only at its cheapest acceptance for $T\ge6$. Both
variants of Snakes and Ladders are slower still, $1.4$--$2.2\times$ the vLLM
kernel for Activation Replay and $2.7$--$3.2\times$ for Joint Attainment,
so they are left out of the figure. Joint Attainment is the faster of the
two at batch size one in its paper, but here the second row per request
doubles the state bytes of the verify launch and lengthens its serial
chain. ReplaySSM is $3.3$--$3.9\times$ faster than STree and
$2.5$--$3.6\times$ faster than The Mamba in the Llama when every token is
committed, and $3.5$--$5.3\times$ and $2.3$--$5.1\times$ with one token
committed.

The prior papers report their speedups at small batch sizes against
conventional standard decoding. At small batch the decode step is dominated
by loading the model weights, and verifying $T$ tokens per step amortizes
that load over several tokens. This gain belongs to speculative decoding
itself and does not depend on how the SSM state is handled. At a serving
batch size the weights are already amortized and the state traffic is the
bottleneck (Section~\ref{sec:challenge-state-io}), so the cost of state
recovery decides whether speculation pays off, and there none of the prior
methods pays off against vLLM's snapshots.

Table~\ref{tab:app:state-traffic} also shows that fewer bytes do not buy
these methods lower latency. At $T\ge4$ STree and The Mamba in the Llama
move $1.6$--$2.9\times$ fewer state bytes than vLLM's kernel, yet they are
not faster. Their recovery kernels are serial recurrences that hold the
state tile in registers for $T$ or $2T-1$ steps, which limits occupancy and
leaves most of the DRAM bandwidth idle, more so as $T$ grows.

\section{GDN derivations and algorithms}
\label{app:gdn}

\subsection{Outer-product and unrolled forms}
\label{app:gdn:derivation}

We derive Equation~\ref{eq:gdn-rewrite} from Equation~\ref{eq:gdn-recurrence}
and unroll it over a replay window, the form the GDN algorithms in this
appendix use.

\paragraph{Outer-product form.}
Distributing the product in Equation~\ref{eq:gdn-recurrence} gives
\begin{align}
    S_t
    &= \alpha_t S_{t-1}\left(I-\beta_t k_t k_t^\top\right)
       + \beta_t v_t k_t^\top \nonumber\\
    &= \alpha_t S_{t-1}
       - \alpha_t\beta_t\left(S_{t-1}k_t\right)k_t^\top
       + \beta_t v_t k_t^\top \nonumber\\
    &= \alpha_t S_{t-1}
       + \beta_t\left(v_t-\alpha_t S_{t-1}k_t\right)k_t^\top .
    \label{eq:app:gdn-expand}
\end{align}
The second line uses $S_{t-1}k_tk_t^\top=(S_{t-1}k_t)k_t^\top$, and the
third collects the two rank-one terms that share the factor $k_t^\top$.
Setting $u_t=\beta_t\left(v_t-\alpha_t S_{t-1}k_t\right)\in\mathbb{R}^{d_v}$
recovers Equation~\ref{eq:gdn-rewrite}. Here $\alpha_t S_{t-1}k_t$ is the
value the decayed state returns for the key $k_t$, so $u_t$ is the part of
$v_t$ that the state does not already hold, scaled by the write strength
$\beta_t$.

ReplaySSM caches $u_t$ in the replay-buffer record, so this
state-dependent computation runs once (Appendix~\ref{app:gdn:standard});
the two-precision variant also keeps $v_t$
(Appendix~\ref{app:gdn:precision}).

\paragraph{Unrolled form.}
Index steps within a replay window by $t=1,2,\ldots$, with
$S_0=S_{\mathrm{ckpt}}$ the checkpoint. Substituting each $S_{t-1}$ into the
update for $S_t$, as in Equation~\ref{eq:mamba2-unrolled}, gives
\begin{equation}
    S_t = \alpha_t S_{t-1} + u_t k_t^\top
    = \gamma_t S_{\mathrm{ckpt}}
    + \sum_{j=1}^{t}\gamma_{j+1:t}\,u_jk_j^\top,
    \qquad
    u_j=\beta_j\left(v_j-\alpha_j S_{j-1}k_j\right).
    \label{eq:app:gdn-unrolled}
\end{equation}
Each $u_j$ depends on $S_{j-1}$, which is Equation~\ref{eq:app:gdn-unrolled}
at $j-1$, so the corrections are determined by $S_{\mathrm{ckpt}}$ and the
window's inputs alone. Once they are cached, the right-hand side at $t=h$ is
$\textsc{Replay}(S_{\mathrm{ckpt}},\mathcal{B}_{1:h})$ from
Equation~\ref{eq:replay} over the records $\{(\alpha_j,u_j,k_j)\}_{j=1}^{h}$,
and it matches the result of $h$ conventional GDN steps up to
floating-point rounding.

\subsection{Standard decoding}
\label{app:gdn:standard}

\begin{algorithm}[!t]
    \caption{ReplaySSM standard decoding step for GDN}
    \label{alg:gdn-standard}
    \algtext*{EndIf}
    \begin{algorithmic}[1]
        \Statex \textbf{Input:} current token's $(\alpha,\beta,v,k)$ and $q$
        \State $S\gets\textsc{Replay}(S_{\mathrm{ckpt}},\mathcal{B}_{1:h})$
            \Comment{state before this token}
        \State $u\gets\beta\left(v-\alpha Sk\right)$
            \Comment{Equation~\ref{eq:gdn-rewrite}}
        \State $\mathcal{B}_{h+1}\gets(\alpha,u,k)$; $h\gets h+1$
        \State $S\gets\alpha S+uk^\top$
        \State $o\gets Sq$
        \If{$h=L$} \Comment{flush}
            \State $S_{\mathrm{ckpt}}\gets S$; store $S_{\mathrm{ckpt}}$ to HBM
            \State $\mathcal{B}\gets\varnothing$; $h\gets0$
        \EndIf
        \State \Return $o$
    \end{algorithmic}
\end{algorithm}

Algorithm~\ref{alg:gdn-standard} is the GDN counterpart of
Algorithm~\ref{alg:replayssm-standard}. The two differ in the order of the
first steps. Mamba-2 appends the record and then reconstructs the state,
because $v_t$ is known before the step. GDN reconstructs $S_{t-1}$ first
(Line~1), since the correction in Line~2 needs $S_{t-1}k_t$, and only then
appends $(\alpha_t,u_t,k_t)$ and advances the state (Lines~3--4). By
Equation~\ref{eq:app:gdn-unrolled}, Line~4 equals
$\textsc{Replay}(S_{\mathrm{ckpt}},\mathcal{B}_{1:h})$ over the updated
buffer, so the flush branch is unchanged.

The correction $u_t$ is computed in FP32 from the reconstructed state. With
BF16 activations we cache it in FP16 rather than BF16, since FP16 keeps
10 mantissa bits of that result against 7 for BF16 at the same storage cost.

\subsection{Speculative decoding}
\label{app:gdn:speculative}

GDN verifies drafts with the output-only evaluation of
Section~\ref{sec:output-only-verification} and keeps the buffer handling of
Algorithm~\ref{alg:replayssm-speculative}: draft records are written behind
the committed ones and the accepted prefix is committed by moving $h$. The
difficulty is that a draft's correction depends on the state just before it
(Equation~\ref{eq:gdn-rewrite}), which includes the effect of the earlier
drafts in the window. Output-only verification still applies: the $T$ draft
corrections follow from one $T\times T$ triangular solve, after which
Equation~\ref{eq:output-only} applies with corrections in place of values.
Algorithm~\ref{alg:gdn-speculative} assembles the step.

Let draft $s$ sit at window position $t=h+s$.
Equation~\ref{eq:app:gdn-unrolled} at $t-1$ gives
\begin{equation}
    \alpha_t S_{t-1}k_t
    = \gamma_t S_{\mathrm{ckpt}}k_t
    + \sum_{j=1}^{t-1}\gamma_{j+1:t}\,(k_j^\top k_t)\,u_j .
    \label{eq:app:gdn-draft-read}
\end{equation}
The terms with $j\le h$ use cached corrections; those with $h<j<t$ involve
earlier drafts, whose corrections are unknown. Let $U_c$, $K_c$ hold the
committed corrections and keys, $V_d$, $K_d$ the draft values and keys, and
$B_d=\operatorname{diag}(\beta_{h+1},\ldots,\beta_{h+T})$. Substituting
Equation~\ref{eq:app:gdn-draft-read} into $u_t=\beta_t(v_t-\alpha_tS_{t-1}k_t)$
for all $T$ drafts and collecting the unknowns as the columns of
$U_d\in\mathbb{R}^{d_v\times T}$ gives
\begin{equation}
    U_d\left(I+AB_d\right)=RB_d,
    \qquad
    R=V_d-S_{\mathrm{ckpt}}K_dD-U_c\bigl(W_c\odot(K_c^\top K_d)\bigr),
    \label{eq:app:gdn-ut}
\end{equation}
where $A_{js}=\gamma_{h+j+1:h+s}\,k_{h+j}^\top k_{h+s}$ for $j<s$ and $0$
otherwise, $D$ is the diagonal decay matrix of
Equation~\ref{eq:output-only}, and $W_c\in\mathbb{R}^{h\times T}$ is the
first $h$ rows of the $(h+T)\times T$ matrix $W$ of
Equation~\ref{eq:output-only}. The
known part $R$ has the shape of Equation~\ref{eq:output-only} with the draft
keys in place of the queries. $I+AB_d$ is unit upper triangular, so one
$T\times T$ triangular solve yields $U_d$; this is the UT transform used to
parallelize delta-rule training \citep{yang2024parallelizing}. With
$U=[U_c,U_d]$, the outputs are Equation~\ref{eq:output-only} with $U$ in
place of $V$:
\begin{equation}
    O=S_{\mathrm{ckpt}}QD+U\bigl(W\odot(K^\top Q)\bigr).
    \label{eq:app:gdn-output-only}
\end{equation}
$S_{\mathrm{ckpt}}$ enters only through $S_{\mathrm{ckpt}}K_d$ and
$S_{\mathrm{ckpt}}Q$, which one pass over the checkpoint computes together.

\begin{algorithm}[h]
    \caption{ReplaySSM speculative verification step for GDN}
    \label{alg:gdn-speculative}
    \algtext*{EndIf}
    \begin{algorithmic}[1]
        \Statex \textbf{Input:} $X=\{(\alpha_s,\beta_s,v_s,k_s)\}_{s=1}^{T}$
            and $Q$ of $T$ tokens
        \State Load $S_{\mathrm{ckpt}}$ from HBM into $S$
        \State $U_d\gets$ solve Equation~\ref{eq:app:gdn-ut} from $S$,
            $\mathcal{B}_{1:h}$, and $X$ \Comment{draft corrections}
        \State $\mathcal{B}_{h+1:h+T}\gets\{(\alpha_s,u_s,k_s)\}_{s=1}^{T}$
            \Comment{$h$ unchanged}
        \State $O\gets\textsc{OutputOnly}(S,\mathcal{B}_{1:h+T},Q)$
            \Comment{Equation~\ref{eq:app:gdn-output-only}}
        \If{$h+T>L$} \Comment{flush committed records}
            \State $S\gets\textsc{Replay}(S,\mathcal{B}_{1:h})$
            \State $S_{\mathrm{ckpt}}\gets S$; store $S_{\mathrm{ckpt}}$ to HBM
            \State Drop $\mathcal{B}_{1:h}$; $h\gets0$ \Comment{keep $\mathcal{B}_{h+1:h+T}$}
        \EndIf
        \State \Return $O$
        \Statex Later, given accepted length $a$:
        \State $h\gets h+a$ \Comment{commit accepted drafts}
    \end{algorithmic}
\end{algorithm}

Algorithm~\ref{alg:gdn-speculative} differs from
Algorithm~\ref{alg:replayssm-speculative} only in its first three lines.
The draft records hold corrections rather than values, so they can be
appended only after Equation~\ref{eq:app:gdn-ut} is solved (Lines~2--3),
which in turn needs the checkpoint loaded first (Line~1). Line~4 is
Equation~\ref{eq:app:gdn-output-only}. Flush, acceptance, and commit are
unchanged: each accepted correction depends only on the checkpoint and the
records before it, so it remains a valid record, and the flush replays
committed records only. In our kernel, Lines~2, 4, and~6 share one pass
over the checkpoint.

\subsection{Low-precision state}
\label{app:gdn:precision}

\begin{algorithm}[!t]
    \caption{Two-precision speculative verification step for GDN}
    \label{alg:gdn-low-precision}
    \definecolor{algchg}{RGB}{0,84,170}
    \providecommand{\chg}[1]{\textcolor{algchg}{#1}}
    \algtext*{EndIf}
    \begin{algorithmic}[1]
        \Statex \textbf{Input:} $X=\{(\alpha_s,\beta_s,v_s,k_s)\}_{s=1}^{T}$
            and $Q$ of $T$ tokens
        \State \chg{$\pi\gets\mathrm{HP}$ if $h+T>L$, else $\mathrm{LP}$}
        \State \chg{Load $S_{\mathrm{ckpt}}^{\pi}$ from HBM into $S$}
        \State $U_d\gets$ solve Equation~\ref{eq:app:gdn-ut} from $S$,
            $\mathcal{B}_{1:h}$, and $X$ \Comment{draft corrections}
        \State \chg{$\mathcal{B}_{h+1:h+T}\gets\{(\alpha_s,\beta_s,u_s,v_s,k_s)\}_{s=1}^{T}$}
            \Comment{$h$ unchanged}
        \State $O\gets\textsc{OutputOnly}(S,\mathcal{B}_{1:h+T},Q)$
            \Comment{Equation~\ref{eq:app:gdn-output-only}}
        \If{$h+T>L$} \Comment{flush committed records}
            \State \chg{Recompute $u_{1:h}$ from $S$ and
                $(\alpha,\beta,v,k)_{1:h}$; $S\gets\textsc{Replay}(S,\mathcal{B}_{1:h})$}
            \State \chg{$S_{\mathrm{ckpt}}^{\mathrm{HP}}\gets\mathcal{Q}_{\mathrm{HP}}(S)$;
                $S_{\mathrm{ckpt}}^{\mathrm{LP}}\gets\mathcal{Q}_{\mathrm{LP}}(S)$}
            \State \chg{Store $S_{\mathrm{ckpt}}^{\mathrm{HP}}$ and
                $S_{\mathrm{ckpt}}^{\mathrm{LP}}$ to HBM}
            \State Drop $\mathcal{B}_{1:h}$; $h\gets0$ \Comment{keep $\mathcal{B}_{h+1:h+T}$}
        \EndIf
        \State \Return $O$
        \Statex Later, given accepted length $a$:
        \State $h\gets h+a$ \Comment{commit accepted drafts}
    \end{algorithmic}
\end{algorithm}

Algorithm~\ref{alg:replayssm-low-precision} keeps low-precision error out of
the recurrence because Mamba-2 records hold raw inputs: a flush replays
$(\alpha,v,k)$ records from the high-precision checkpoint, and nothing read
from the low-precision copy is written back. GDN records hold corrections,
and each correction is computed from the state its step read
(Equation~\ref{eq:gdn-rewrite}). On non-flush steps that state comes from
the low-precision checkpoint, so the cached $u_j$ carry its rounding error.
If a flush folds these corrections into the high-precision checkpoint, the
error enters the recurrence, and each flush cycle adds its own: the
checkpoint error becomes a sum over all past cycles, finite for any decay
below one but growing without bound as the decay approaches one. Folding
cached corrections is therefore stable only under a data-dependent
condition.

We avoid this by caching the raw inputs as well. In the two-precision
variant a GDN record holds $(\alpha,\beta,u,v,k)$. Non-flush steps use $u$
as before. A flush ignores the cached corrections and re-derives them from
the high-precision checkpoint along Equation~\ref{eq:app:gdn-unrolled},
$u_j=\beta_j(v_j-\alpha_jS_{j-1}k_j)$ with $S_{j-1}$ on the high-precision
path. The low-precision copy then never enters the recurrence, and the
high-precision path is exact up to its own rounding, as for Mamba-2. The
two flush paths differ only when the read checkpoint is low precision. A
cached correction carries the rounding of the state it was computed from,
so folding it writes that error into the checkpoint, where every later
window inherits it. Re-deriving it from the high-precision checkpoint and
the cached raw inputs folds a clean value, so the error stays in the
outputs computed before the flush. With an FP32 checkpoint the sole
difference is that a cached correction is stored in the FP16 record format
while a re-derived one is formed in FP32 on the fly. Our FP32 speedup
measurements therefore fold the cached corrections, and all low-precision
results re-derive them.

Algorithm~\ref{alg:gdn-low-precision} extends
Algorithm~\ref{alg:gdn-speculative} the way
Algorithm~\ref{alg:replayssm-low-precision} extends
Algorithm~\ref{alg:replayssm-speculative}. The changes, shown in color, are
the precision choice and load (Lines~1--2), the extended record (Line~4),
the re-derivation at flush (Line~7), and the two casts and stores
(Lines~8--9). Acceptance, rollback, and buffer management are unchanged.

\clearpage
\subsection{Notation for the algorithms}
\label{app:notation}

Table~\ref{tab:notation} defines the symbols of
Algorithms~\ref{alg:replayssm-standard}--\ref{alg:gdn-low-precision}. The
last column states how GDN departs from Mamba-2; ``same'' means the symbol
is unchanged.

\begin{table}[h]
    \caption{Symbols in Algorithms~\ref{alg:replayssm-standard}--\ref{alg:gdn-low-precision}.}
    \label{tab:notation}
    \centering
    \small
    \renewcommand{\arraystretch}{1.1}
    \setlength{\tabcolsep}{5pt}
    \begin{tabular}{@{}l>{\raggedright\arraybackslash}p{0.38\linewidth}>{\raggedright\arraybackslash}p{0.29\linewidth}@{}}
        \toprule
        Symbol & Meaning (Mamba-2) & GDN \\
        \midrule
        $S_{\mathrm{ckpt}}$ &
            Checkpoint state in HBM; the recurrent state at the end of the
            last flushed window. &
            same \\
        \midrule
        $S_{\mathrm{ckpt}}^{\mathrm{HP}}$, $S_{\mathrm{ckpt}}^{\mathrm{LP}}$ &
            High- and low-precision copies of the checkpoint. HP carries the
            recurrence across flushes; LP serves non-flush output computation. &
            same \\
        \midrule
        $S$ &
            Working state on chip; written to HBM only at a flush, as the new
            checkpoint. &
            same \\
        \midrule
        $x=(\alpha,v,k)$, $q$ &
            SSM inputs of one token: decay, value, key, and query
            (Eq.~\ref{eq:mamba2-recurrence}). &
            Inputs are $(\alpha,\beta,v,k)$ with write strength $\beta$
            (Eq.~\ref{eq:gdn-recurrence}). Records store the correction $u$
            in place of $v$; two-precision records are $(\alpha,\beta,u,v,k)$. \\
        \midrule
        $u$, $U_c$, $U_d$ &
            -- &
            Correction $\beta(v-\alpha Sk)$ (Eq.~\ref{eq:gdn-rewrite});
            committed and draft corrections as matrix columns
            (Eq.~\ref{eq:app:gdn-ut}). \\
        \midrule
        $\mathcal{B}$, $\mathcal{B}_{i:j}$ &
            Replay buffer and its records at positions $i$ to $j$; position
            $1$ is the oldest record since the last flush. &
            same \\
        \midrule
        $h$ &
            Number of committed records $\mathcal{B}_{1:h}$. Standard
            decoding commits on append; speculative decoding commits drafts
            after verification. &
            same \\
        \midrule
        $L$ &
            Replay window size, the maximum committed history before a
            flush: $h=L$ in standard decoding, $h+T>L$ in speculative
            decoding. &
            same \\
        \midrule
        $T$, $X$, $Q$ &
            Number of tokens per verification step, $1\le T\le L$;
            their records $X=\{x_s\}_{s=1}^{T}$ and queries
            $Q=[q_1,\ldots,q_T]$. &
            $x_s$ includes $\beta_s$. \\
        \midrule
        $a$ &
            Number of tokens committed after verification, $1\le a\le T$. &
            same \\
        \midrule
        $\textsc{Replay}(S,\mathcal{B}_{1:h})$ &
            State after applying records $1$ to $h$ to $S$
            (Eq.~\ref{eq:replay}). &
            Records supply $u_j$ (Eq.~\ref{eq:app:gdn-unrolled}); a
            two-precision flush first recomputes $u_j$ from
            $(\beta_j,v_j,k_j)$ on the HP path. \\
        \midrule
        $\textsc{OutputOnly}(S,\mathcal{B}_{1:h+T},Q)$ &
            The $T$ outputs for $Q$ from $S$ and the records, without
            intermediate states (Eq.~\ref{eq:output-only}). &
            Eq.~\ref{eq:app:gdn-output-only} with $U$ in place of $V$, after
            solving Eq.~\ref{eq:app:gdn-ut} for $U_d$. \\
        \midrule
        $\pi$ &
            Checkpoint precision loaded this step: HP on flush steps, LP
            otherwise. &
            same \\
        \midrule
        $\mathcal{Q}_{\mathrm{HP}}$, $\mathcal{Q}_{\mathrm{LP}}$ &
            Casts from accumulation precision to the two checkpoint formats. &
            same \\
        \bottomrule
    \end{tabular}
\end{table}
\FloatBarrier

\section{Cost and numerical error analysis}
\label{app:analysis}

\subsection{Arithmetic intensity of conventional decoding}
\label{app:analysis:intensity}

Taking Mamba-2 as the example, we count the bytes moved and the FLOPs
performed by one head in one conventional decoding step
(Equation~\ref{eq:mamba2-recurrence}). The state
$S_t\in\mathbb{R}^{d_v\times d_k}$ is stored with $b$ bytes per element
($b=4$ for FP32, $b=2$ for FP16 or BF16).

\paragraph{Memory traffic.}
The step loads $S_{t-1}$ and stores $S_t$, moving $2b\,d_vd_k$ bytes. It
also reads $\alpha_t$, $v_t$, $k_t$, and $q_t$ and writes $o_t$, but these
are vectors of length $d_v$ or $d_k$, negligible against the $d_vd_k$
state.

\paragraph{FLOPs.}
A multiplication and an addition count as two FLOPs. The decay
$\alpha_tS_{t-1}$, the outer product $v_tk_t^\top$, their sum, and the
readout $S_tq_t$ cost $d_vd_k$, $d_vd_k$, $d_vd_k$, and $2d_vd_k-d_v$ FLOPs,
so the step performs $5d_vd_k-d_v$ FLOPs.

\paragraph{Arithmetic intensity.}
The ratio of FLOPs to bytes is
\begin{equation}
    I_{\mathrm{conv}}
    = \frac{5d_vd_k-d_v}{2b\,d_vd_k}
    \approx \frac{5}{2b}
    \;\text{FLOP per byte},
    \label{eq:app:intensity}
\end{equation}
about $0.63$ for an FP32 state and $1.25$ for FP16. Batching does not
raise it, because every sequence brings its own state and no state element
is reused, so traffic and FLOPs scale together. An H100 needs about $20$
FLOP per byte of HBM traffic to keep its FP32 CUDA cores busy, and about
$300$ for its BF16 tensor cores \citep{nvidia2022h100}. Conventional SSM
decoding runs at under $1$, so the step is bottlenecked by moving the state
through HBM, not by arithmetic.

\subsection{SSM and attention traffic in a hybrid model}
\label{app:analysis:crossover}

Section~\ref{sec:challenge-state-io} states that the SSM layers of
Nemotron-3-Super-120B-A12B-NVFP4 move more bytes per decoding step than its
attention layers until the context reaches about 80K tokens. We count the
bytes here from the shapes in the released model
configuration.\footnote{\url{https://huggingface.co/nvidia/NVIDIA-Nemotron-3-Super-120B-A12B-NVFP4/blob/main/config.json}}
Table~\ref{tab:exp:model-geometry} lists the SSM shapes.

\paragraph{State size.}
Each Mamba-2 layer has 128 value heads with $d_v=64$ and $d_k=128$ and
stores its state in FP32, so the state is $128\times64\times128\times4$
bytes, or 4~MiB per sequence. A decoding step loads and stores it, moving
8~MiB per sequence per layer.

\paragraph{KV cache size.}
Each attention layer has 2 key-value heads of dimension 128 and stores keys
and values in FP8, so one token's entry is $2\times2\times128$ bytes, or
0.5~KiB. One SSM state therefore equals the KV cache for
$4~\mathrm{MiB}/0.5~\mathrm{KiB}=8192$ tokens. A decoding step at context
length $N$ loads the $N$ entries, about $0.5N$~KiB per sequence per layer.

\paragraph{Crossover.}
With 40 Mamba-2 layers and 8 attention layers, the SSM layers move
$40\times8=320$~MiB per sequence per step and the attention layers move
$8\times0.5N=4N$~KiB. The totals are equal at $N=81{,}920$ tokens, the
layer ratio of five times 8K tokens times two for the load and the store.
Batching does not move the crossover, since both totals scale with the
number of sequences. The measured crossover in
Figure~\ref{fig:replayssm-overview} (bottom left) comes later, near 100K
tokens, which is consistent with attention kernels having received more
optimization effort \citep[e.g.,][]{dao2022flashattention}.

\subsection{Memory traffic}
\label{app:analysis:traffic}

We compare the HBM traffic of conventional decoding and ReplaySSM standard
decoding (Algorithm~\ref{alg:replayssm-standard}) for one Mamba-2 head over
one replay window of $L$ steps, with an FP32 state, BF16 values and keys,
and an FP32 decay. The window has $L-1$ non-flush steps and one flush step.
Both methods read the current token's $\alpha_t$, $v_t$, $k_t$, and $q_t$
and write $o_t$ at every step; this traffic is identical on both sides and
negligible (Appendix~\ref{app:analysis:intensity}), so we omit it from both
totals. The buffer traffic that ReplaySSM adds is counted.

\paragraph{Traffic per window.}
Conventional decoding loads and stores the $4d_vd_k$-byte state at every
step, so one window moves $8L\,d_vd_k$ bytes. ReplaySSM loads the
checkpoint at every step and stores it once, at the flush, $4(L+1)\,d_vd_k$
bytes, and adds the buffer traffic. A record holds $v_j$ and $k_j$ in BF16
and $\alpha_j$ in FP32, $r=2d_v+2d_k+4$ bytes. The $i$-th non-flush step,
$i=1,\ldots,L-1$, loads the $i-1$ buffered records and stores the current
input as one record; the flush step loads all $L-1$ records and folds the
current input into the checkpoint directly, storing no record. Over the
window, reads total $\sum_{i=1}^{L-1}(i-1)+(L-1)=L(L-1)/2$ records and
writes total $L-1$, so $(L-1)(L+2)/2$ record transfers in all.
The two totals are
\begin{equation}
    M_{\mathrm{conv}}=8L\,d_vd_k,
    \qquad
    M_{\mathrm{replay}}=4(L+1)\,d_vd_k+\frac{(L-1)(L+2)}{2}\,r .
    \label{eq:app:traffic}
\end{equation}
The checkpoint term alone gives the ratio $(L+1)/(2L)$, just above one half
for any $L$; the buffer term grows quadratically in $L$ and sets the
window-size tradeoff of Appendix~\ref{app:analysis:window}.

\paragraph{Nemotron-3-Super-120B-A12B.}
Its Mamba-2 layers have $d_v=64$ and $d_k=128$. With $L=16$,
Equation~\ref{eq:app:traffic} gives
$M_{\mathrm{replay}}/M_{\mathrm{conv}}=0.58$, a $1.7\times$ reduction in
memory traffic. The figure is conservative. Mamba-2 shares each key across
the $16$ heads of a group, so the per-head key traffic is smaller than the
record size counts.

\paragraph{Speculative decoding.}
Table~\ref{tab:app:state-traffic} in Appendix~\ref{app:prior-methods}
extends this count to verification steps and compares it with vLLM's
snapshots and the prior recovery methods, as read and write counts and as
Nsight bytes.

\subsection{FLOPs for output-only computation}
\label{app:analysis:flops}

Section~\ref{sec:output-only-verification} introduced output-only
evaluation, which computes $o=v(k^\top q)$ without materializing the state.
Because $k$ and $q$ are shared by the heads of a group, $k^\top q$ is also
computed once per group rather than once per head. Here we count the FLOPs.

FLOPs do not set the latency of a memory-bound kernel in standard decoding
with an FP32 state (Appendix~\ref{app:analysis:intensity}), but they matter
as arithmetic intensity rises. A lower-precision state moves fewer bytes per
step, and speculative decoding multiplies the arithmetic per step by the
number of tokens verified. Not materializing the state also saves on-chip
memory. We count per head with the rule of
Appendix~\ref{app:analysis:intensity} and keep only the leading term of
each count in $d_v$ and $d_k$.

\paragraph{Standard decoding.}
Algorithm~\ref{alg:replayssm-standard} materializes the state before the
readout, and the main text presents only that route for standard decoding.
Output-only evaluation applies there as well, so we count both routes. With
$h$ buffered records and the current input, a step applies $h+1$
value-key pairs. The state-and-output route
(Algorithm~\ref{alg:replayssm-standard}) costs
\begin{itemize}
    \item decayed checkpoint plus $h+1$ weighted outer products, summed:
        $2(h+2)\,d_vd_k$;
    \item readout $S_tq_t$: $2d_vd_k$;
\end{itemize}
about $2(h+3)\,d_vd_k$ in total. The output-only route evaluates
\begin{equation}
    o_t=\gamma_{h+1}S_{\mathrm{ckpt}}q_t
    +\sum_{j=1}^{h+1}\gamma_{j+1:h+1}\,v_j\,(k_j^\top q_t)
    \label{eq:app:output-only-standard}
\end{equation}
and costs
\begin{itemize}
    \item checkpoint readout $S_{\mathrm{ckpt}}q_t$: $2d_vd_k$;
    \item $h+1$ inner products $k_j^\top q_t$: $2(h+1)\,d_k$, once per group;
    \item weighted sum of the $h+1$ values: $2(h+1)\,d_v$;
\end{itemize}
about $2d_vd_k+2(h+1)(d_v+d_k)$ in total. Both routes share the checkpoint
readout. The term that materializes the state, $2(h+1)\,d_vd_k$, becomes
$2(h+1)(d_v+d_k)$, smaller by a factor $d_vd_k/(d_v+d_k)$, about $43$ for
the shapes of Appendix~\ref{app:analysis:traffic}.

\paragraph{Speculative decoding.}
Equation~\ref{eq:output-only} verifies $T$ tokens against $h$ committed
records. Its three terms cost
\begin{itemize}
    \item checkpoint readout $S_{\mathrm{ckpt}}QD$ against $T$ queries:
        $2T\,d_vd_k$;
    \item $T(h+T)$ inner products in $K^\top Q$: $2T(h+T)\,d_k$, once per
        group;
    \item weighted sum $V(W\odot K^\top Q)$: $2T(h+T)\,d_v$;
\end{itemize}
about $2T\,d_vd_k+2T(h+T)(d_v+d_k)$ in total, which per token is the
standard-decoding count with $h+1$ replaced by $h+T$. Materializing states
instead would replay the committed state, about $2(h+2)\,d_vd_k$, and then
run $T$ conventional steps of $5d_vd_k$ each
(Appendix~\ref{app:analysis:intensity}), about $(2h+4+5T)\,d_vd_k$. For
$T=4$, a full window of $h=12$, and the shapes above, output-only
verification needs over $4\times$ fewer FLOPs. Its only remaining
$d_vd_k$-scale term is the single readout of the checkpoint against the $T$
queries, and $K^\top Q$ becomes one small GEMM per group.

\subsection{Replay-window tradeoff}
\label{app:analysis:window}

A larger window amortizes each checkpoint write over more tokens but makes
every step load more records. We extend the traffic analysis of
Appendix~\ref{app:analysis:traffic} to speculative decoding and solve for
the window that minimizes traffic per committed token, treating the kernel
as purely memory-bound. In practice the FLOPs and on-chip buffer storage of
Appendix~\ref{app:analysis:flops} also enter, so the result is an
approximation. Let
$M_S=4d_vd_k$ be the state size in bytes, $r$ the record size, $T$ the
tokens per step, and $\bar a$ the mean number of tokens committed per
step. One flush cycle commits $L$ tokens over $L/\bar a$ steps and moves
\begin{itemize}
    \item checkpoint reads, one per step: $M_S\,L/\bar a$;
    \item checkpoint write, one per cycle: $M_S$;
    \item buffer reads: at the $i$-th step of the cycle the buffer holds
        $i\bar a$ committed records, $i=0,\ldots,L/\bar a-1$, so
        $r\bar a\sum_i i=rL(L-\bar a)/(2\bar a)$;
    \item buffer writes: $T$ records per step, written before verification
        so rejected drafts are paid for too, and none at the flush step:
        $rT(L/\bar a-1)$.
\end{itemize}
Dividing by $L$ gives the traffic per committed token,
\begin{equation}
    f(L)
    = \frac{M_S}{\bar a}+\frac{M_S}{L}
    + \frac{rL}{2\bar a}-\frac{r}{2}
    + \frac{rT}{\bar a}-\frac{rT}{L},
    \label{eq:app:window-model}
\end{equation}
which for $\bar a=T=1$ equals $M_{\mathrm{replay}}/L$ of
Equation~\ref{eq:app:traffic}. Setting $f'(L)=0$ gives
\begin{equation}
    L^{*}=\sqrt{2\bar a\left(\frac{M_S}{r}-T\right)}.
    \label{eq:app:window-opt}
\end{equation}
The optimum grows with the state-to-record size ratio and with the square
root of the acceptance rate, since higher acceptance commits more tokens per
checkpoint read. The model also treats a cycle as exactly $L/\bar a$ steps.
The flush fires when $h+T>L$, so a cycle commits between $L-T+1$ and $L$
tokens depending on where the last acceptance lands, and flushes come
somewhat more often than the model assumes.

\paragraph{Nemotron-3-Super-120B-A12B.}
With the shapes of Appendix~\ref{app:analysis:traffic}, $M_S/r=84$.
Standard decoding ($\bar a=T=1$) gives $L^{*}\approx13$, and speculative
decoding with $T=4$ and $\bar a=3.5$ gives $L^{*}\approx24$. The sweep in
Appendix~\ref{app:exp:window} finds $L=16$ best among $8$, $16$, and $32$
for all four models, the sampled value nearest the standard-decoding
prediction. Accounting for the key shared within a group would raise
$L^{*}$, but FLOPs and on-chip storage also grow with $L$.

\subsection{Quantization error propagation}
\label{app:analysis:error}

\paragraph{Rounding modes.}
Storing the state in a low-precision format rounds each element to one of
its two representable neighbours, which lie a distance $\delta$ apart that
depends on the element's magnitude. Round-to-nearest (RTN) picks the closer
one. Its error is at most $\delta/2$, but its sign depends on the value, so
the errors of similar values share a direction and add up. Stochastic
rounding (SR) uses the position of the value within the gap. If the value
lies a fraction $p$ of the way from the lower neighbour to the upper one, SR
rounds up with probability $p$ and down with probability $1-p$, so the
rounded value equals the original in expectation. SR removes the bias but
not the error. Each rounding is a zero-mean error with variance
$p(1-p)\delta^2$, at most $\delta^2/4$. The analysis below models these two
behaviours as a per-element bias $b$ and variance $\sigma^2$.

\paragraph{Propagation through the recurrence.}
Let $S_t$ follow Equation~\ref{eq:mamba2-recurrence} exactly, and let
$\hat S_t$ be the stored state, which is rounded by $\mathcal{Q}$ after
every update: $\hat S_t=\mathcal{Q}(X_t)$ with
$X_t=\alpha_t\hat S_{t-1}+v_tk_t^\top$. The rounding error of step $t$ is
$e_t=\mathcal{Q}(X_t)-X_t$, and the accumulated deviation of the stored
state from the exact one is $D_t=\hat S_t-S_t$. Subtracting the two
recurrences gives
\begin{equation}
    D_t=\alpha_tD_{t-1}+e_t
    =\sum_{j\le t}\gamma_{j+1:t}\,e_j,
    \label{eq:app:error-recurrence}
\end{equation}
so each step's rounding error is decayed by later steps but never removed,
and the output error is $D_tq_t$. To see the scaling, take a constant decay
$\alpha$ and errors that are independent across steps with bias $b$ and
variance $\sigma^2$ per element. The mean deviation converges to
$b/(1-\alpha)$ and the variance to $\sigma^2/(1-\alpha^2)$, both reached
after a number of steps on the order of $1/(1-\alpha)$. RTN contributes
through the first term and SR only through the second. Both limits are
finite for $\alpha<1$ and grow without bound as $\alpha\to1$. Since
$\alpha$ is a per-head, data-dependent quantity, the saturation level is
head dependent. Heads with slow decay, which hold information over long
horizons, accumulate the most error under this analysis.

\paragraph{Effect of ReplaySSM.}
With a single low-precision checkpoint, rounding occurs once per window
rather than once per step, and the deviation decays by $\gamma_L=\alpha^L$
between injections. The limits become $b/(1-\alpha^L)$ and
$\sigma^2/(1-\alpha^{2L})$, lower than the conventional ones because
$\alpha^L<\alpha$, but still growing as $\alpha\to1$. With separate
precisions, the recurrence runs on the high-precision checkpoint and the
low-precision copy is read only for outputs, so its rounding error never
enters Equation~\ref{eq:app:error-recurrence}. The output error at a
non-flush step is a single rounding, independent of $\alpha$ and of the
sequence length.

\paragraph{Limits of the analysis.}
The analysis predicts how error scales with decay and window size, not its
magnitude. It assumes independent errors with fixed bias and variance,
whereas real rounding errors depend on the values being rounded and can
partly cancel. We therefore measure the error directly as
the KL divergence between the next-token distributions of low-precision and
FP32-state decoding; Appendix~\ref{app:exp:kl} defines the measurement and
Section~\ref{sec:exp-low-precision} reports it.

\section{Serving implementation}
\label{app:implementation}

We implement ReplaySSM in vLLM for Mamba-2 and
GDN, but nothing in the design depends on vLLM. The released source code
records the buffer layout, its placement in the paged state, and the
cursor bookkeeping. This appendix covers the two decisions that determine
how the kernels behave inside a serving engine, the flush rule under
speculative decoding and execution under CUDA graphs.

\subsection{Flush rule under speculative decoding}
\label{app:implementation:flush}

In standard decoding the buffer holds $L$ records, each token is committed
when it is appended, and the flush fires when the buffer is full
(Algorithm~\ref{alg:replayssm-standard}). Speculative decoding appends $T$
records per step and commits only $a$ of them, and $a$ is known only after
the step, so the rule changes. The ring has capacity $L+T$, and we keep the
invariant that at most $L$ committed records are in the buffer at the start
of a step, so the $T$ records of the step always fit behind them. Suppose a
step with $h$ committed records does not flush and all $T$ tokens are
accepted. The committed length becomes $h+T$, and the next step appends $T$
more records at positions up to $h+2T$. When $h+T>L$ this exceeds the
capacity $L+T$, and in a ring the new records overwrite committed ones.
Algorithm~\ref{alg:replayssm-speculative} therefore flushes exactly when
$h+T>L$. The flush folds the $h$ committed records into the checkpoint
before acceptance is known, which is valid because they stay committed
whatever the verifier returns, and the committed length after the step is
$a\le T\le L$, so the invariant holds.

The rule depends only on $h$ and $T$, both known before the step runs, so
the decision never waits on sampling. It is an early flush in the sense that
it may fire when a partial acceptance would have left room. The flush
interval is therefore not fixed at $L$ steps as in standard decoding but
roughly $L/a$ steps, and higher acceptance flushes more often. This is the
band in Figures~\ref{fig:exp:speculative-kernel-latency-mamba2}
and~\ref{fig:exp:speculative-kernel-latency-gdn}.

\subsection{CUDA graphs}
\label{app:implementation:graphs}

Serving engines capture the decode step as a CUDA graph and replay it, which
removes the CPU launch overhead of the many small kernels in a step. A
captured graph fixes the launch sequence for a batch size, so any decision
that varies within a batch has to be made on the device. ReplaySSM needs two
such decisions.

The first is whether a row flushes. Under continuous batching the rows of a
batch were admitted at different times and sit at different buffer
positions, so they reach their flush steps at different times. Speculative
decoding adds a second source of divergence, because each row commits its
own $a$. At a given step some rows flush and others do not, and one graph
has to serve them all. ReplaySSM therefore treats the flush decision as
per-row data that the kernels read at run time rather than a compile-time
constant. The flush work is either a branch inside the decode kernel or a
separate launch specialized for flushing, from which non-flushing rows exit
at once. Either way one captured graph per batch size covers both kinds of
rows, and the inactive rows of a padded batch exit the same way.

The second is the commit after verification. The accepted count $a$ is
known only after sampling and differs across rows. Reading it back to the
host would add a device-to-host synchronization at every step and stall the
pipeline. Instead a small commit kernel runs on the device. It advances
each row's buffer pointer by that row's $a$, moves the ring origin past the
flushed records on a flush step, and sets the flush flags of the next step,
since the rule of Appendix~\ref{app:implementation:flush} depends only on
the resulting $h$ and on $T$. Every cursor update stays inside the captured
work.

\section{Additional evaluation details and results}
\label{app:evaluation}

\subsection{Evaluation setup}
\label{app:exp:setup}

\begin{table}[!t]
    \caption{Layer counts and SSM-layer shapes. Each key head serves the
    value heads of its group.}
    \label{tab:exp:model-geometry}
    \centering
    \footnotesize
    \setlength{\tabcolsep}{6pt}
    \begin{tabular}{@{}llccccc@{}}
        \toprule
        Model & Weights & SSM / Attn. layers & Key heads & Value heads & $d_k$ & $d_v$ \\
        \midrule
        Nemotron-3-Nano-4B & BF16 & 21 / 4 & 8 & 96 & 128 & 80 \\
        Nemotron-3-Super-120B-A12B & NVFP4 & 40 / 8 & 8 & 128 & 128 & 64 \\
        Qwen3.5-4B & BF16 & 24 / 8 & 16 & 32 & 128 & 128 \\
        Qwen3.5-122B-A10B & NVFP4 & 36 / 12 & 16 & 64 & 128 & 128 \\
        \bottomrule
    \end{tabular}
\end{table}

\paragraph{Models.}
Table~\ref{tab:exp:model-geometry} adds the layer counts and the SSM-layer
shape of each model to Table~\ref{tab:exp:model-settings}; the kernel
benchmarks in this appendix run at these shapes. Nemotron-3-Super-120B-A12B
and both Qwen3.5 models ship a multi-token-prediction (MTP) head; the
speculative-decoding experiments use it as the drafter on the two
120B-class models. Throughout the paper, ``FP32 state'' refers to the SSM
state only; weights keep their released precision
(Table~\ref{tab:exp:model-settings}) and activations are BF16. FP32 is also the SSM state precision all
four models default to in vLLM. The Qwen3.5 models set
\texttt{mamba\_ssm\_dtype} and Nemotron-3-Super-120B-A12B sets
\texttt{mamba\_ssm\_cache\_dtype} to \texttt{float32} in their released
configurations, and Nemotron-3-Nano-4B leaves it unset, so vLLM applies its
\texttt{float32} default for NemotronH models.\footnote{Configuration files
of \url{https://huggingface.co/Qwen/Qwen3.5-4B},
\url{https://huggingface.co/nvidia/Qwen3.5-122B-A10B-NVFP4}, and
\url{https://huggingface.co/nvidia/NVIDIA-Nemotron-3-Super-120B-A12B-NVFP4};
the NemotronH default is set in
\texttt{vllm/model\_executor/models/config.py} of the vLLM repository.}

\paragraph{Hardware and software.}
The 4B models run on one H100 with 80\,GB of HBM3, and the 120B-class
models on one B300 under a 1100\,W power cap, all without tensor
parallelism. The performance experiments run the ReplaySSM kernels inside
vLLM against vLLM's own kernels, both in the same build with CUDA graphs
enabled. The low-precision experiments use emulated quantization. The KL
and accuracy measurements run a Triton kernel that implements
Algorithm~\ref{alg:replayssm-low-precision} with quantized checkpoints
inside the same vLLM integration, and the low-precision latency
measurements (Appendix~\ref{app:exp:precision-latency}) use separate CuTe
DSL kernels.

\paragraph{Measurement protocol.}
Unless a figure states otherwise, performance measurements use an FP32
state, $L=16$, and greedy decoding. Kernel latencies are standalone
measurements at the model's SSM shape, after warmup and with a cold L2, and
each ReplaySSM measurement covers full replay windows. Conventional kernels
are timed at the best launch configuration found by a sweep
(Appendix~\ref{app:exp:batch}). Reported latencies are medians over
repeated launches.

\subsection{Standard decoding}
\label{app:exp:standard}

\begin{table}[!t]
    \caption{End-to-end per-step speedup over tuned conventional vLLM
    standard decoding ($L=16$, FP32 state). $^\dagger$Inflated by the B300
    power cap, which throttles the conventional kernel inside the full model
    but not ReplaySSM; the kernel speedups in
    Table~\ref{tab:exp:standard-batch-kernel} are unaffected.}
    \label{tab:exp:standard-batch-e2e}
    \centering
    \footnotesize
    \setlength{\tabcolsep}{4pt}
\begin{tabular}{@{}llcccccc@{}}
\toprule
Model & Weights & $B\!=\!1$ & $B\!=\!8$ & $B\!=\!32$ & $B\!=\!128$ & $B\!=\!256$ & $B\!=\!512$ \\
\midrule
Nemotron-3-Nano-4B & BF16 & $0.98\times$ & $1.03\times$ & $1.16\times$ & $1.38\times$ & $1.47\times$ & $1.51\times$ \\
Nemotron-3-Super-120B-A12B & NVFP4 & $0.96\times$ & $1.01\times$ & $1.06\times$ & $1.24\times$ & $1.42\times$ & $1.72\times^\dagger$ \\
Qwen3.5-4B & BF16 & $0.99\times$ & $1.03\times$ & $1.08\times$ & $1.21\times$ & $1.27\times$ & $1.30\times$ \\
Qwen3.5-122B-A10B & NVFP4 & $0.96\times$ & $1.03\times$ & $1.09\times$ & $1.16\times$ & $1.20\times$ & $1.22\times^\dagger$ \\
\bottomrule
\end{tabular}

    \vspace{10pt}
    \captionof{table}{Isolated SSM-kernel speedup over tuned conventional vLLM
    standard decoding ($L=16$, FP32 state).}
    \label{tab:exp:standard-batch-kernel}
\begin{tabular}{@{}llcccccc@{}}
\toprule
Model & Weights & $B\!=\!1$ & $B\!=\!8$ & $B\!=\!32$ & $B\!=\!128$ & $B\!=\!256$ & $B\!=\!512$ \\
\midrule
Nemotron-3-Nano-4B & BF16 & $0.73\times$ & $1.36\times$ & $1.69\times$ & $1.78\times$ & $1.80\times$ & $1.80\times$ \\
Nemotron-3-Super-120B-A12B & NVFP4 & $0.41\times$ & $1.02\times$ & $1.20\times$ & $1.46\times$ & $1.53\times$ & $1.56\times$ \\
Qwen3.5-4B & BF16 & $0.88\times$ & $1.34\times$ & $1.54\times$ & $1.66\times$ & $1.68\times$ & $1.67\times$ \\
Qwen3.5-122B-A10B & NVFP4 & $0.51\times$ & $1.16\times$ & $1.37\times$ & $1.44\times$ & $1.46\times$ & $1.49\times$ \\
\bottomrule
\end{tabular}

\end{table}

\subsubsection{Batch-size sweep}
\label{app:exp:batch}

\begin{figure}[!t]
    \centering
    \includegraphics[width=\linewidth]{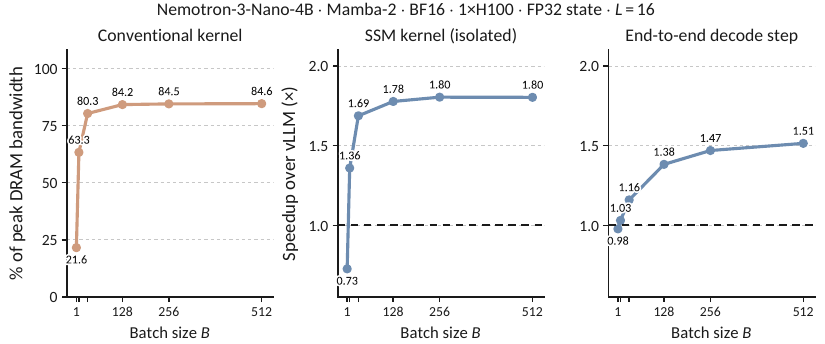}
    \caption{Standard decoding across batch size on Nemotron-3-Nano-4B
    ($L=16$, FP32 state, one H100). Left: DRAM bandwidth of the
    conventional kernel as a fraction of peak. Middle and right: isolated
    kernel and end-to-end per-step speedup over tuned conventional vLLM.}
    \label{fig:exp:standard-batch-sweep}
\end{figure}

The baseline is vLLM's kernel at the best launch configuration found by a
sweep, not at its shipped default. vLLM picks its Mamba-2 configuration
from a heuristic with no tuned entry for either shape we evaluate, while
the shipped GDN configuration is already the swept
optimum.\footnote{Against the shipped Mamba-2 default, the speedups at
$B=256$ would read $1.49\times$ end-to-end and $1.89\times$ for the kernel
on Nemotron-3-Nano-4B, and $1.48\times$ and $1.75\times$ on
Nemotron-3-Super-120B-A12B, instead of the $1.47$/$1.80$ and $1.42$/$1.53$
we report.} ReplaySSM speeds up standard decoding at serving batch sizes, where the
conventional kernel is memory-bound, and not at small batches, where it is
not. Figure~\ref{fig:exp:standard-batch-sweep} sweeps the batch size on
Nemotron-3-Nano-4B, and Tables~\ref{tab:exp:standard-batch-e2e}
and~\ref{tab:exp:standard-batch-kernel} give the end-to-end and kernel
speedups for all four models. At $B=1$ the speedup is below parity
($0.41$--$0.88\times$ for the kernel, $0.96$--$0.99\times$ for the decode
step), because the conventional kernel reaches only $21.6\%$ of peak DRAM
bandwidth on H100 and $9.3\%$ on B300, so reducing its memory traffic saves
nothing.
From $B\approx128$ the baseline saturates at $84$--$86\%$ of peak and the
kernel speedup levels off at $1.4$--$1.8\times$, while the end-to-end
speedup keeps growing because the SSM layers take a larger share of the
decode step as the batch grows. The end-to-end gain is also largest at short
and medium sequence lengths; at long sequences attention dominates the step,
since its KV-cache traffic grows linearly with sequence length
(Figure~\ref{fig:replayssm-overview}, bottom left).

\begin{figure}[!t]
    \centering
    \includegraphics[width=\linewidth]{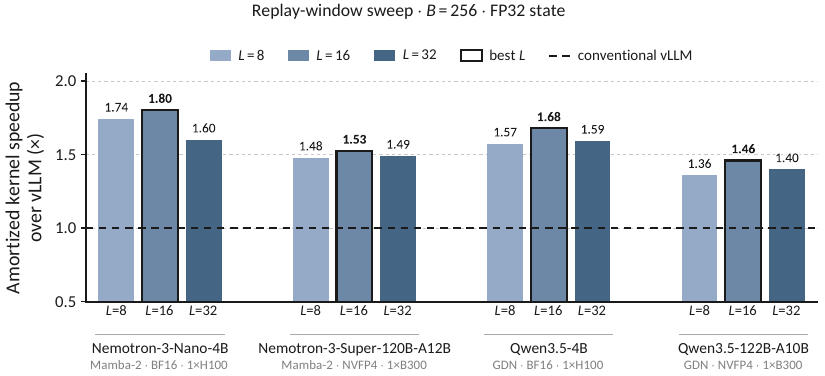}
    \caption{Amortized kernel speedup over tuned conventional vLLM across
    replay-window sizes, amortized over one replay cycle ($B=256$, FP32
    state). $L=16$ is best in all four settings. The vertical axis starts at
    $0.5$.}
    \label{fig:exp:standard-window-sweep}
\end{figure}

\subsubsection{Replay-window sweep}
\label{app:exp:window}

The window size trades flush frequency against per-step work. A shorter
window flushes more often and pays for more checkpoint writes; a longer
window flushes less often, but every step reads more from the buffer.
Figure~\ref{fig:exp:standard-window-sweep} sweeps $L\in\{8,16,32\}$ at
$B=256$. The amortized kernel latency is lowest at $L=16$ for all four
models. The traffic analysis of Appendix~\ref{app:analysis:window} predicts an
optimum near $L=13$ for standard decoding, and $L=16$ is the nearest power
of two.

\FloatBarrier
\subsection{Speculative decoding}
\label{app:exp:speculative}

Both models draft with their MTP head. The acceptance length is the mean
number of tokens committed per step. Requests are GSM8K prompts
\citep{gsm8k} served at a fixed batch size $B$; every request generates
256 tokens with end-of-sequence ignored, and the timed window covers the
whole request, prefill included, after warmup. Throughput is the total
number of generated tokens divided by the wall time.

\subsubsection{Serving throughput across batch size and verification length}
\label{app:exp:serving}

Tables~\ref{tab:exp:speculative-speedup-mamba2}
and~\ref{tab:exp:speculative-speedup-gdn} give the throughput of ReplaySSM
speculative decoding relative to conventional vLLM speculative decoding
over the full grid of $B$ and $T$, and
Table~\ref{tab:exp:speculative-acceptance} the acceptance lengths at
$B=256$. Three points hold on both models. At $B=1$ the two methods are
close ($0.88$--$1.17\times$ on Nemotron-3-Super-120B-A12B-NVFP4,
$0.87$--$1.12\times$ on Qwen3.5-122B-A10B-NVFP4), because at small batch the
state traffic that ReplaySSM removes is not the bottleneck. At larger
batch the speedup grows with both $B$ and $T$, and it has two sources.
Conventional speculation stores one state snapshot per draft token, so its
state traffic and storage grow with $B$ and $T$ together; the storage caps
the concurrency that GPU memory allows, and its throughput saturates from
$B\approx128$ and at large $T$ falls below standard
decoding.\footnote{On Nemotron-3-Super-120B-A12B-NVFP4 at $B=256$, conventional
speculation peaks at $T=2$ with 9{,}551 tokens/s and drops to 5{,}644 at
$T=8$, below standard decoding at 6{,}807.} ReplaySSM writes the state
only on flush steps and verifies with the faster kernel of
Figure~\ref{fig:exp:speculative-kernel-latency-mamba2}, so it keeps
scaling. Acceptance lengths agree within $0.05$ tokens in every cell
(Table~\ref{tab:exp:speculative-acceptance}), so the gain does not come
from a different acceptance behavior.

\begin{table}[!t]
    \begin{minipage}[t]{0.485\linewidth}
        \centering
        \begin{minipage}[t][5\baselineskip][t]{\linewidth}
            \caption{ReplaySSM speculative decoding over conventional
            speculative decoding, Nemotron-3-Super-120B-A12B-NVFP4, $L=16$.}
            \label{tab:exp:speculative-speedup-mamba2}
        \end{minipage}
        \footnotesize
        \setlength{\tabcolsep}{2.5pt}
\begin{tabular}{@{}lccccc@{}}
\toprule
 & \multicolumn{5}{c}{Batch size $B$} \\
\cmidrule(lr){2-6}
$T$ & $1$ & $32$ & $128$ & $256$ & $512$ \\
\midrule
$2$ & $0.88\times$ & $1.15\times$ & $1.24\times$ & $1.34\times$ & $1.78\times$ \\
$4$ & $1.04\times$ & $1.13\times$ & $1.29\times$ & $1.77\times$ & $1.85\times$ \\
$6$ & $1.05\times$ & $1.03\times$ & $1.48\times$ & $1.91\times$ & $2.11\times$ \\
$8$ & $1.17\times$ & $1.38\times$ & $1.90\times$ & $2.09\times$ & $2.34\times$ \\
\bottomrule
\end{tabular}

    \end{minipage}\hfill
    \begin{minipage}[t]{0.485\linewidth}
        \centering
        \begin{minipage}[t][5\baselineskip][t]{\linewidth}
            \caption{ReplaySSM speculative decoding over conventional
            speculative decoding, Qwen3.5-122B-A10B-NVFP4, $L=16$.}
            \label{tab:exp:speculative-speedup-gdn}
        \end{minipage}
        \footnotesize
        \setlength{\tabcolsep}{2.5pt}
\begin{tabular}{@{}lccccc@{}}
\toprule
 & \multicolumn{5}{c}{Batch size $B$} \\
\cmidrule(lr){2-6}
$T$ & $1$ & $32$ & $128$ & $256$ & $512$ \\
\midrule
$2$ & $1.00\times$ & $1.05\times$ & $1.35\times$ & $1.36\times$ & $1.73\times$ \\
$4$ & $1.12\times$ & $1.22\times$ & $1.43\times$ & $1.79\times$ & $2.10\times$ \\
$6$ & $0.87\times$ & $1.17\times$ & $1.42\times$ & $1.99\times$ & OOM \\
$8$ & $0.96\times$ & $1.12\times$ & $1.84\times$ & $2.09\times$ & OOM \\
\bottomrule
\end{tabular}

    \end{minipage}
\end{table}

\begin{table}[!t]
    \caption{Mean acceptance length at $B=256$.}
    \label{tab:exp:speculative-acceptance}
    \centering
    \footnotesize
    \setlength{\tabcolsep}{5pt}
\begin{tabular}{@{}lcccc@{}}
\toprule
 & \multicolumn{2}{c}{Nemotron-3-Super-120B-A12B-NVFP4} & \multicolumn{2}{c}{Qwen3.5-122B-A10B-NVFP4} \\
\cmidrule(lr){2-3}\cmidrule(lr){4-5}
 & Conventional & ReplaySSM & Conventional & ReplaySSM \\
\midrule
$T\!=\!2$ & $1.95$ & $1.95$ & $1.95$ & $1.95$ \\
$T\!=\!4$ & $3.43$ & $3.44$ & $3.56$ & $3.57$ \\
$T\!=\!6$ & $4.24$ & $4.20$ & $4.73$ & $4.76$ \\
$T\!=\!8$ & $4.40$ & $4.44$ & $5.52$ & $5.53$ \\
\bottomrule
\end{tabular}

\end{table}

\subsubsection{Kernel latency on GDN}
\label{app:exp:spec-kernel}

\begin{figure}[!tb]
    \centering
    \includegraphics[width=\linewidth]{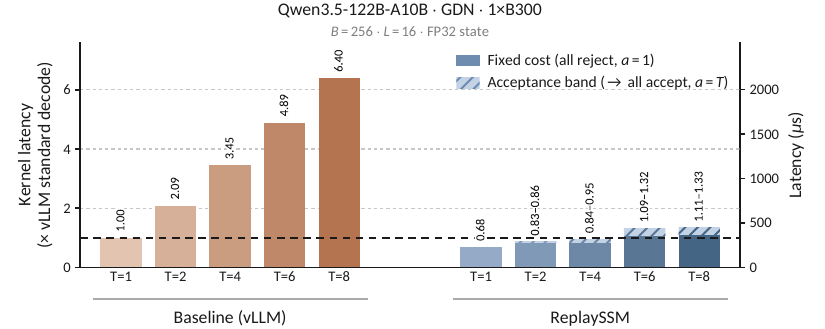}
    \caption{Speculative kernel latency on Qwen3.5-122B-A10B (GDN; $B=256$,
    $L=16$, FP32 state), normalized to conventional standard decoding
    ($T=1$). ReplaySSM bands span all drafts rejected to all drafts
    accepted and include flush costs.}
    \label{fig:exp:speculative-kernel-latency-gdn}
\end{figure}

Figure~\ref{fig:exp:speculative-kernel-latency-gdn} repeats
Figure~\ref{fig:exp:speculative-kernel-latency-mamba2} for
Qwen3.5-122B-A10B. GDN shows
the same trend as Mamba-2. The conventional kernel costs $2.09\times$ a
standard decoding step at $T=2$ and $6.40\times$ at $T=8$, while ReplaySSM
stays at $0.83$--$0.95\times$ up to $T=4$ and $1.11$--$1.33\times$ at
$T=8$, flushes included, with each band spanning $a=1$ to $a=T$
(Appendix~\ref{app:implementation:flush}). The step between $T=4$ and
$T=6$ comes from tiling. Some kernel internals pad the verification window
to a multiple of four tokens, so $T=6$ pays for eight.

\subsection{Low-precision state}
\label{app:exp:precision}

\subsubsection{Formats and quantization}
\label{app:exp:formats}

We store checkpoints in four formats. FP32 and FP16 are unscaled. FP8 uses
E4M3 elements with one FP32 scale per block of 128 contiguous elements along
the key dimension, which is one scale per state row since $d_k=128$ for all
four models. NVFP4 uses E2M1 elements with one E4M3 scale per block of 16
elements and one FP32 scale per state. A flush accumulates the new
checkpoint in FP32 from the recurrence checkpoint and the buffered inputs and
casts that FP32 result into each stored copy independently, never one copy
from another. The cast into the recurrence checkpoint uses stochastic
rounding when that checkpoint is FP16 or FP8, and the cast into the read
checkpoint uses round-to-nearest. A shared checkpoint is a single copy
written with stochastic rounding; Appendix~\ref{app:exp:rounding} shows
what happens when it is written with round-to-nearest instead.
Quantization applies to decoding only. Prefill
computes the state at full precision, and rounding enters only through
checkpoint writes during generation. All low-precision results use $L=16$
and standard decoding. Speculative decoding changes how many tokens a step
verifies, and acceptance sets when flushes occur. Neither changes which
values are rounded, so the error mechanism is the same.

\subsubsection{Latency of low-precision reads}
\label{app:exp:precision-latency}

\begin{figure}[!t]
    \centering
    \includegraphics[width=\linewidth]{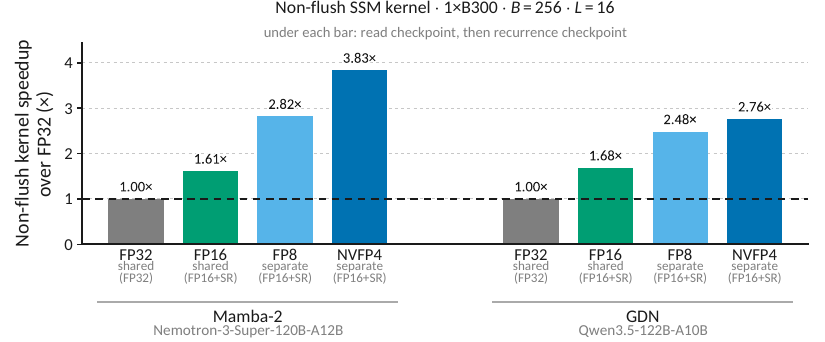}
    \caption{Non-flush kernel speedup over an FP32 state by read-checkpoint
    precision ($B=256$, $L=16$, one B300).}
    \label{fig:exp:precision-nonflush-speedup}
\end{figure}

\begin{figure}[!t]
    \centering
    \includegraphics[width=\linewidth]{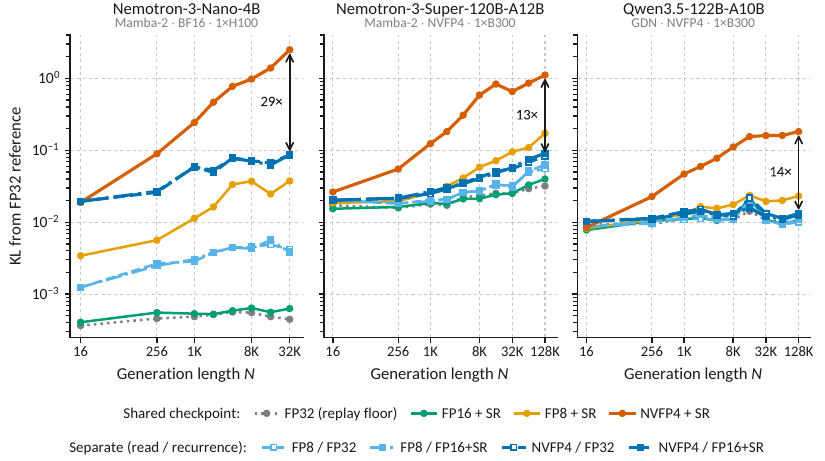}
    \caption{Output KL from conventional FP32-state decoding under teacher
    forcing on PG-19 ($L=16$), the three models outside
    Figure~\ref{fig:exp:precision-kl-small-models}, on one shared vertical
    axis. Nemotron-3-Nano-4B stops at 32K (see text).
    Arrows mark the gap between a shared NVFP4 checkpoint and NVFP4 reads
    with FP32 recurrence at the last plotted $N$.}
    \label{fig:exp:precision-kl-by-model}
\end{figure}

The point of the separate checkpoints in
Section~\ref{sec:replayssm-low-precision} is to keep rounding error out of
the recurrence. A lower-precision read checkpoint also moves fewer bytes on
every non-flush step, so it can lower latency as well, and
Figure~\ref{fig:exp:precision-nonflush-speedup} quantifies that gain. It
measures the non-flush kernel of
standard decoding at $B=256$ with CuTe DSL kernels, varying only the read
checkpoint while the FP16+SR recurrence checkpoint, which a non-flush step
never touches, stays fixed. Relative to an FP32 state, an FP16 read runs
$1.61\times$ faster on Mamba-2 and $1.68\times$ on GDN, an FP8 read
$2.82\times$ and $2.48\times$, and an NVFP4 read $3.83\times$ and
$2.76\times$. These are non-flush ratios. Over a full window the gain is
smaller, since flush steps still read and write the higher-precision
checkpoint; on Mamba-2 at $B=512$ the NVFP4 read gives $4.11\times$ on
non-flush steps and $3.33\times$ over the window. Under speculative
decoding it is smaller still, since flushes are more frequent and each step
carries more arithmetic.

\subsubsection{Output KL across models}
\label{app:exp:kl}

\paragraph{Setting.}
We decode four streams, each one book from the PG-19 test split
\citep{rae2019compressive}, from a 512-token prompt under teacher forcing. At
every step we record the model's logits and replace the sampled token with the
book's next token, so all configurations consume the same token sequence
and only the recurrent state differs. The metric is
$D_{\mathrm{KL}}(p_{\mathrm{FP32}}\Vert p_{\mathrm{config}})$ over the full
vocabulary, averaged over the four streams and the 256 positions ending at
$N$, where $p_{\mathrm{FP32}}$ is conventional FP32-state decoding with
CUDA graphs. One seed per configuration.

\paragraph{Results.}
Figure~\ref{fig:exp:precision-kl-by-model} shows the three models not in
Figure~\ref{fig:exp:precision-kl-small-models}, and three points hold
across all four. First, a shared low-precision checkpoint accumulates
error with $N$, while a separate recurrence removes most of the growth. At
the last plotted $N$, NVFP4 reads with FP32 recurrence sit $13$--$29\times$
below a shared NVFP4 checkpoint (arrows in
Figure~\ref{fig:exp:precision-kl-by-model}), and FP16+SR recurrence tracks
FP32 recurrence within $1.25\times$ at every point but one. The growth is
reduced rather than removed, since on the two Nemotron models the curves
with a separate recurrence still rise a few times over the range. Second,
a lower read precision raises the starting point of a curve. On
Nemotron-3-Nano-4B at $N=16$, an FP8 read sits at $1.2\times10^{-3}$ and an
NVFP4 read at $2\times10^{-2}$, against a floor of $3.7\times10^{-4}$.
Third, Nemotron-3-Super-120B-A12B-NVFP4 and Qwen3.5-122B-A10B-NVFP4 have
floors about two orders of magnitude above the 4B models, at
$1$--$3\times10^{-2}$. They differ from the 4B models in size, weight
precision, and expert routing at once, and our data do not attribute the
floor to any one of these. Two conventional FP32 runs that differ only in
execution path, eager against CUDA graph, are already this far apart on
these models, so the floor is numerical noise rather than error that
ReplaySSM introduces, and each configuration should be judged by its
distance above the floor of its panel. Nemotron-3-Nano-4B is plotted only
to 32K, because beyond that its FP32 reference becomes nearly one-hot on
the end-of-turn token, which deflates every configuration's
KL.\footnote{The mean probability of the end-of-turn token is $0.76$ at
64K and $0.88$ at 128K. The other three models place no measurable
probability on that token at any $N$.}

\subsubsection{Window-size sensitivity of KL}
\label{app:exp:kl-window}

\begin{figure}[!t]
    \centering
    \includegraphics[width=\linewidth]{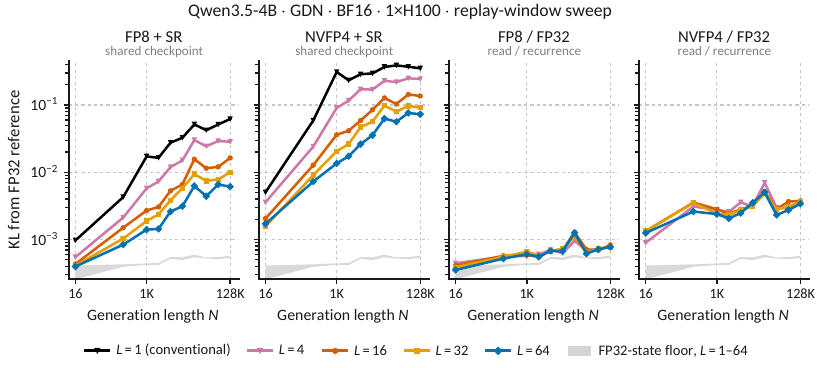}
    \caption{Output KL across replay-window sizes on Qwen3.5-4B. $L=1$
    requantizes the state at every step, as conventional decoding does, and
    exists only for a shared checkpoint. Separate configurations use FP32
    recurrence. The grey band is the FP32-state ReplaySSM floor over
    $L=1$--$64$.}
    \label{fig:exp:precision-kl-window-sweep}
\end{figure}

Figure~\ref{fig:exp:precision-kl-window-sweep} repeats the KL measurement
on Qwen3.5-4B at $L\in\{1,4,16,32,64\}$ for two shared and two separate
configurations; the separate ones use FP32 recurrence here. $L=1$ is
conventional decoding, where the state is written and downcast at every
step. It exists only for a shared checkpoint, since the separate
configurations need the non-flush steps that a window of $L>1$ creates.

The two kinds of configuration respond to $L$ in the two ways
Section~\ref{sec:replayssm-low-precision} describes. A shared checkpoint is
requantized at every flush, so a longer window means fewer rounding events
and a lower KL. From $L=1$ to $64$ the number of checkpoint writes falls
$64\times$. Over the same range the KL of the shared FP8 checkpoint falls
$12\times$ at $N=1$K and $10\times$ at 128K, and that of the shared NVFP4
checkpoint falls $23\times$ and $4.8\times$. The gain is less than
proportional to the number of rounding events, as
Appendix~\ref{app:analysis:error} predicts. At the $L=16$ we use, the
shared FP8 and NVFP4 checkpoints sit $3.8\times$ and $2.6\times$ below
conventional per-step quantization at 128K, which is the first benefit of
ReplaySSM for a low-precision state. The separate configurations barely
change across $L$. They move by $1.1$--$1.2\times$ between $L=4$ and $64$,
the same as the scatter of the FP32-state floor itself
($1.05$--$1.06\times$). Their read checkpoint is regenerated from the
unquantized recurrence at every flush and never fed back, so each output
sees one rounding of a clean state however many flushes preceded it, and
$L$ barely affects it.\footnote{On GDN the corrections $u$ cached on
non-flush steps are computed from the low-precision read checkpoint and
enter the outputs of the rest of the window, which leaves a small
dependence on $L$. The flush re-derives them from the high-precision path
(Appendix~\ref{app:gdn:precision}), so this error does not reach the
recurrence.}

\subsubsection{Stochastic rounding against round-to-nearest}
\label{app:exp:rounding}

Appendix~\ref{app:exp:formats} writes the recurrence checkpoint with
stochastic rounding and the read copy with round-to-nearest.
Table~\ref{tab:exp:precision-rounding} gives the reason. It compares the two
rounding modes on a shared checkpoint at $L=16$ through the ratio of the RTN
KL to the SR KL at matched precision, which cancels the FP32 floor. At short
generation lengths RTN is close to or below SR. At $N=1$K it gives
$0.40$--$0.51\times$ the SR error for NVFP4 and $0.59$--$1.05\times$ for
FP8 across the four models, because SR's random error is present at full
size from the first rounding and grows only in quadrature, while RTN's
bias is small per step but adds linearly and overtakes it once enough steps
have accumulated (Appendix~\ref{app:analysis:error}). The ratio then rises
with $N$ and
crosses one by 16K tokens for FP8 on every model, and on the two GDN models RTN drives the FP8 and NVFP4 state
non-finite by 128K and 64K tokens while SR stays finite everywhere. FP16,
not tabulated, shows the same crossover on the two Mamba-2 models, where RTN
reaches
$15\times$ SR on Nemotron-3-Nano-4B at 32K and $2.7\times$ on
Nemotron-3-Super-120B-A12B at 128K, and stays within $2.3\times$ on the GDN
models. A checkpoint that is read back into the recurrence therefore needs
SR, and a copy that is rounded once and never fed back is better served by
RTN.

\begin{table}[!t]
    \caption{Output KL of a shared checkpoint with round-to-nearest divided
    by that with stochastic rounding, at matched precision and $L=16$.
    Above one, RTN is worse. div.: the RTN state became non-finite before
    this $N$. Nemotron-3-Nano-4B is read to 32K
    (Appendix~\ref{app:exp:kl}).}
    \label{tab:exp:precision-rounding}
    \centering
    \footnotesize
    \setlength{\tabcolsep}{5pt}
\begin{tabular}{@{}llrrrr@{}}
\toprule
 & & \multicolumn{4}{c}{Generation length $N$} \\
\cmidrule(lr){3-6}
Model & Precision & 1K & 8K & 32K & 128K \\
\midrule
Nemotron-3-Nano-4B & FP8 & 0.59 & 1.3 & 6.6 & -- \\
 & NVFP4 & 0.51 & 0.67 & 0.90 & -- \\
\midrule
Nemotron-3-Super-120B-A12B & FP8 & 1.1 & 1.4 & 2.0 & 2.2 \\
 & NVFP4 & 0.44 & 0.45 & 1.4 & 0.95 \\
\midrule
Qwen3.5-4B & FP8 & 0.76 & 1.2 & 2.2 & div. \\
 & NVFP4 & 0.40 & 1.0 & 4.1 & div. \\
\midrule
Qwen3.5-122B-A10B & FP8 & 0.94 & 0.85 & 1.7 & div. \\
 & NVFP4 & 0.50 & 1.4 & 13 & div. \\
\bottomrule
\end{tabular}

\end{table}

\subsubsection{Downstream accuracy}
\label{app:exp:accuracy}

\begin{table}[!t]
    \caption{Downstream accuracy for every model, configuration, and
    benchmark. $\Delta$ is the change from that model's FP32 reference in
    percentage points, and the FP32 row gives its absolute accuracy. Tokens
    are the mean and 90th percentile of output length in thousands.}
    \label{tab:exp:precision-accuracy-full}
    \centering
    \footnotesize
    \setlength{\tabcolsep}{3pt}
\begin{tabular}{@{}lrcrcrc@{}}
\toprule
 & \multicolumn{2}{c}{MMLU-Pro} & \multicolumn{2}{c}{GPQA-Diamond} & \multicolumn{2}{c}{Math-L5} \\
\cmidrule(lr){2-3}\cmidrule(lr){4-5}\cmidrule(lr){6-7}
Configuration & $\Delta$ (pp) & Tokens (K) & $\Delta$ (pp) & Tokens (K) & $\Delta$ (pp) & Tokens (K) \\
\midrule
\multicolumn{7}{@{}l}{\textit{Nemotron-3-Nano-4B}} \\
\addlinespace[1pt]
FP32 accuracy (\%) & $67.2$ & 2.9 / 6.9 & $56.7$ & 10.3 / 26.7 & $89.5$ & 2.1 / 4.5 \\
Shared FP16+SR & $-0.3$ & 2.9 / 6.8 & $-2.2$ & 10.7 / 29.9 & $-1.1$ & 2.2 / 4.9 \\
Shared FP8+SR & $-0.6$ & 3.0 / 6.9 & $-1.0$ & 10.7 / 29.0 & $-1.0$ & 2.2 / 4.9 \\
Shared NVFP4+SR & $-21.0$ & 2.1 / 3.6 & $-31.3$ & 4.3 / 5.0 & $-41.7$ & 1.7 / 3.2 \\
Separate FP8 / FP16+SR & $-0.6$ & 2.9 / 6.7 & $-3.5$ & 10.6 / 28.8 & $-0.6$ & 2.2 / 4.5 \\
Separate NVFP4 / FP16+SR & $-1.7$ & 3.1 / 7.2 & $-6.7$ & 12.0 / 33.0 & $-1.5$ & 2.2 / 4.7 \\
Separate NVFP4 / FP8+SR & $-2.1$ & 3.0 / 7.0 & $-5.6$ & 11.7 / 31.3 & $-2.8$ & 2.3 / 4.9 \\
\midrule
\multicolumn{7}{@{}l}{\textit{Nemotron-3-Super-120B-A12B}} \\
\addlinespace[1pt]
FP32 accuracy (\%) & $81.4$ & 3.7 / 10.4 & $79.3$ & 19.6 / 66.1 & $95.5$ & 4.2 / 9.4 \\
Shared FP16+SR & $+0.3$ & 3.7 / 10.4 & $-3.4$ & 19.5 / 64.4 & $+0.2$ & 4.2 / 9.6 \\
Shared FP8+SR & $+0.2$ & 3.7 / 10.2 & $0.0$ & 19.5 / 68.4 & $+0.7$ & 4.2 / 10.0 \\
Shared NVFP4+SR & $-0.9$ & 3.7 / 10.5 & $-12.5$ & 28.3 / 93.2 & $-2.9$ & 5.0 / 10.2 \\
Separate FP8 / FP16+SR & $+0.2$ & 3.7 / 10.3 & $+0.2$ & 18.8 / 65.5 & $+0.3$ & 4.2 / 10.1 \\
Separate NVFP4 / FP16+SR & $+0.5$ & 3.8 / 10.8 & $-1.4$ & 20.2 / 72.7 & $-0.1$ & 4.5 / 10.1 \\
Separate NVFP4 / FP8+SR & $+0.2$ & 3.7 / 10.5 & $-1.4$ & 19.4 / 64.9 & $+0.1$ & 4.3 / 9.6 \\
\midrule
\multicolumn{7}{@{}l}{\textit{Qwen3.5-4B}} \\
\addlinespace[1pt]
FP32 accuracy (\%) & $79.0$ & 12.5 / 27.4 & $76.3$ & 29.5 / 61.0 & $93.7$ & 15.9 / 31.3 \\
Shared FP16+SR & $-0.5$ & 12.5 / 27.1 & $-1.5$ & 29.9 / 62.2 & $-0.3$ & 15.9 / 32.4 \\
Shared FP8+SR & $-0.4$ & 12.5 / 27.1 & $-0.5$ & 29.5 / 58.0 & $-0.5$ & 16.0 / 31.8 \\
Shared NVFP4+SR & $-1.1$ & 11.8 / 25.9 & $-3.2$ & 27.8 / 60.4 & $-1.7$ & 16.2 / 32.9 \\
Separate FP8 / FP16+SR & $-0.4$ & 12.5 / 27.3 & $+2.4$ & 29.7 / 58.8 & $-0.3$ & 16.2 / 32.5 \\
Separate NVFP4 / FP16+SR & $-0.2$ & 12.4 / 27.1 & $+0.7$ & 29.6 / 58.0 & $-0.1$ & 15.9 / 31.3 \\
Separate NVFP4 / FP8+SR & $-0.6$ & 12.4 / 27.3 & $+1.5$ & 29.4 / 60.6 & $-0.3$ & 16.1 / 32.3 \\
\midrule
\multicolumn{7}{@{}l}{\textit{Qwen3.5-122B-A10B}} \\
\addlinespace[1pt]
FP32 accuracy (\%) & $82.9$ & 8.2 / 17.0 & $82.7$ & 15.2 / 27.8 & $90.7$ & 12.2 / 23.7 \\
Shared FP16+SR & $+0.1$ & 8.1 / 16.7 & $-1.0$ & 15.3 / 28.3 & $+1.2$ & 12.0 / 22.7 \\
Shared FP8+SR & $-0.1$ & 8.3 / 17.1 & $-2.7$ & 15.5 / 29.2 & $+0.9$ & 12.1 / 23.4 \\
Shared NVFP4+SR & $-0.5$ & 7.2 / 13.9 & $-5.2$ & 12.9 / 33.8 & $+2.5$ & 9.8 / 16.4 \\
Separate FP8 / FP16+SR & $+0.6$ & 7.9 / 16.2 & $0.0$ & 15.3 / 28.4 & $+0.1$ & 12.4 / 24.4 \\
Separate NVFP4 / FP16+SR & $+0.8$ & 7.8 / 15.8 & $-2.0$ & 15.5 / 30.4 & $+1.3$ & 12.0 / 22.6 \\
Separate NVFP4 / FP8+SR & $+0.8$ & 7.9 / 16.1 & $-1.7$ & 15.7 / 30.2 & $+1.5$ & 11.9 / 22.6 \\
\bottomrule
\end{tabular}

\end{table}

\paragraph{Protocol.}
MMLU-Pro has 12{,}032 questions, GPQA-Diamond 198 questions with three
repetitions, and Math-L5 1{,}324 questions. Each model uses its
recommended sampling settings. The Nemotron models use temperature $1.0$
and top-$p$ $0.95$; Qwen3.5-4B uses temperature $1.0$, top-$p$ $0.95$,
top-$k$ $20$, and a presence penalty of $1.5$; Qwen3.5-122B-A10B uses
temperature $0.6$, top-$p$ $0.95$, and top-$k$ $20$. The output budget of
each model and benchmark is set near the 99th percentile of the lengths at
which FP32 decoding finishes, and a generation that hits the budget counts
as wrong, since lenient grading would favor the configurations that
truncate most. Every configuration of a model shares the budget, batch
size, sampling settings, and per-question seed, and the change from FP32
is computed on the questions the two runs share, so the comparison is
paired.

\paragraph{Results.}
Table~\ref{tab:exp:precision-accuracy-full} gives the per-model changes
behind Table~\ref{tab:exp:precision-accuracy} together with the mean and
90th-percentile output length of each cell. Shared FP16+SR and FP8+SR
checkpoints stay within benchmark noise of FP32 on every model. The loss
of a shared NVFP4+SR checkpoint follows generation length. On every model,
GPQA-Diamond has the longest generations of the three benchmarks and also
the largest loss, $31.3$ points on Nemotron-3-Nano-4B, $12.5$ on
Nemotron-3-Super-120B-A12B, $3.2$ on Qwen3.5-4B, and $5.2$ on
Qwen3.5-122B-A10B, against at most $2.9$ points on MMLU-Pro and Math-L5
for the three larger models. Nemotron-3-Nano-4B, the smallest model, loses
$21$ to $42$ points on all three. Part of the GPQA-Diamond loss is failure
to finish. On Nemotron-3-Super-120B-A12B, $20.7\%$ of the shared-NVFP4
generations hit the budget against $5.7\%$ for FP32, the loss among pairs
that both finished is $2.8$ points, and the length columns show the same
with a mean of 28.3K tokens and a 90th percentile at the budget. This is
consistent with error that accumulates over a generation
(Section~\ref{sec:challenge-quantization}), and
Appendix~\ref{app:exp:accuracy-length} confirms it within each benchmark,
where length is not confounded with the benchmark. Separate checkpoints
recover most of the loss. With an NVFP4 read and FP16+SR or FP8+SR
recurrence, the three larger models stay within $2$ points of FP32 on
every benchmark, and Nemotron-3-Nano-4B still loses $5.6$ to $6.7$ points
on GPQA-Diamond and $1.5$ to $2.8$ elsewhere. GPQA-Diamond has 594
generations per cell and a question-clustered standard error of $2.5$ to
$3$ points, so only its larger changes are resolved.

\newpage
\paragraph{Long input context.}
\label{app:exp:ruler}
The three benchmarks above test long context on the output side, with
prompts of about 1K tokens and generations that reach tens of thousands.
RULER-128K \citep{hsieh2024ruler} tests it on the input side, with prompts
of about 130K tokens and answers of 18 to 81 tokens. Table~\ref{tab:exp:ruler}
gives the RULER score of every configuration on its 13 tasks with 100
examples each, under RULER's protocol of greedy decoding with thinking
disabled. Every configuration is within $0.3$ points of FP32 on every model,
a shared NVFP4+SR checkpoint included. This is expected. Quantization
applies to decoding only (Section~\ref{sec:exp-low-precision}), so a request
here quantizes its post-prefill state once and flushes at most five times,
against hundreds to nearly two thousand flushes per request on the
reasoning benchmarks. The
accuracy loss of a shared low-precision checkpoint therefore appears with
long outputs, not with long inputs.

\begin{table}[!t]
    \caption{RULER-128K score for every model and configuration. The FP32
    row gives the absolute score and the other rows the change from it in
    percentage points.}
    \label{tab:exp:ruler}
    \centering
    \footnotesize
    \setlength{\tabcolsep}{5pt}
\begin{tabular}{@{}lrrrr@{}}
\toprule
Configuration & \begin{tabular}[b]{@{}r@{}}Nemotron-3-\\Nano-4B\end{tabular} & \begin{tabular}[b]{@{}r@{}}Nemotron-3-Super-\\120B-A12B-NVFP4\end{tabular} & \begin{tabular}[b]{@{}r@{}}Qwen3.5-4B\end{tabular} & \begin{tabular}[b]{@{}r@{}}Qwen3.5-122B-\\A10B-NVFP4\end{tabular} \\
\midrule
FP32 score & $90.4$ & $96.9$ & $95.2$ & $96.8$ \\
Shared FP16+SR & $+0.1$ & $-0.1$ & $+0.1$ & $-0.1$ \\
Shared FP8+SR & $0.0$ & $-0.1$ & $+0.1$ & $0.0$ \\
Shared NVFP4+SR & $-0.3$ & $0.0$ & $0.0$ & $+0.1$ \\
Separate FP8 / FP16+SR & $+0.1$ & $-0.1$ & $+0.2$ & $+0.1$ \\
Separate NVFP4 / FP16+SR & $+0.2$ & $-0.1$ & $+0.1$ & $+0.1$ \\
Separate NVFP4 / FP8+SR & $+0.2$ & $-0.2$ & $+0.1$ & $+0.2$ \\
\bottomrule
\end{tabular}

\end{table}

\subsubsection{Accuracy by generation length}
\label{app:exp:accuracy-length}

\begin{figure}[!t]
    \centering
    \includegraphics[width=\linewidth]{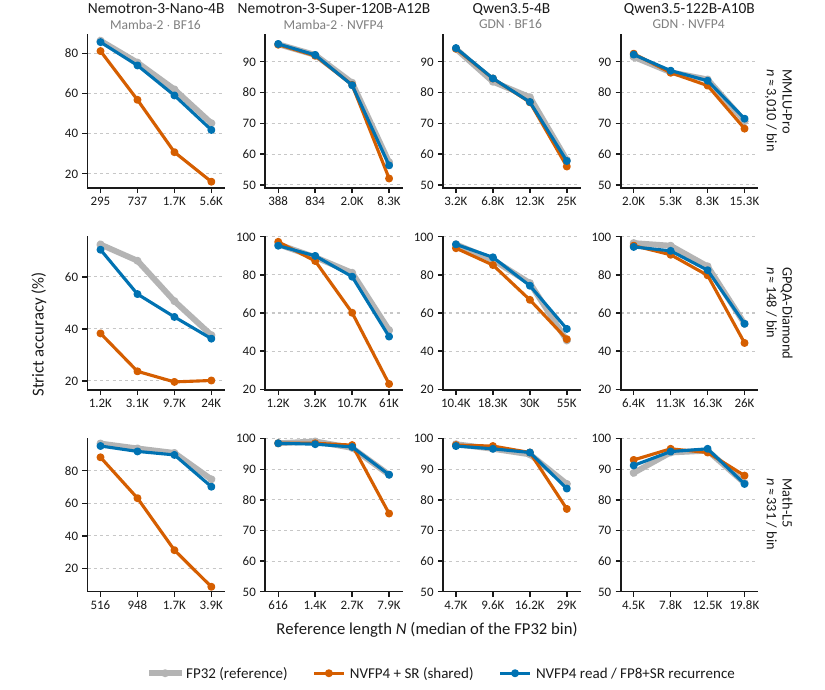}
    \caption{Accuracy by quartile of FP32 reference generation length, all
    models and benchmarks. Bins are defined by the FP32 run. Within each
    benchmark row the three larger models share one vertical range; the
    Math-L5 row starts at 50\%.}
    \label{fig:exp:precision-accuracy-by-length}
\end{figure}

A single accuracy number does not show where a configuration loses.
Figure~\ref{fig:exp:precision-accuracy-by-length} therefore extends
Figure~\ref{fig:exp:precision-gpqa-by-length} to all four models and three
benchmarks and plots accuracy against generation length for the FP32
reference, the shared NVFP4+SR checkpoint, and NVFP4 reads with FP8+SR
recurrence. Within each model and benchmark, the bins are quartiles of FP32
output length per question. Each generation is binned by the length of its
FP32 counterpart, not its own, since a degraded configuration rambles and
would otherwise look long. The FP32 curve comes from a separate FP32 run,
because the run that defines the bins puts all of its own truncated
generations in the top bin. Each bin is plotted at its median reference
length. FP32 accuracy itself falls by $21$ to $50$ points across the bins
on MMLU-Pro and GPQA-Diamond, since longer generations come from harder
questions, so the quantity to read is the gap between a configuration and
FP32 within a bin, not the absolute accuracy.

Wherever the shared NVFP4 checkpoint loses accuracy, its gap to FP32 grows
with length. On the three larger models it matches FP32 in the short bins
and loses in the top bin, by up to $28.2$ points on
Nemotron-3-Super-120B-A12B GPQA-Diamond, and every such top-bin loss comes
with more truncation than FP32. On Nemotron-3-Nano-4B it is below FP32 in
every bin, with a deficit that grows from $5$ to $29$ points on MMLU-Pro
and from $8$ to $66$ on Math-L5. NVFP4 reads with FP8+SR recurrence stay
within the FP32 interval in 41 of the 48 bins.

\FloatBarrier

\section{Future work: skipping checkpoint reads by decay}
\label{app:future-decay-skip}

ReplaySSM removes the state write from non-flush steps but still reads the
checkpoint at every step, because output computation needs it
(Section~\ref{sec:standard-decoding}). The checkpoint enters each step only
through the cumulative decay. On a non-flush step with $h$ buffered records
the output is
\begin{equation}
    o_t=\gamma_{h}\,S_{\mathrm{ckpt}}\,q_t
    +\sum_{j=1}^{h}\gamma_{j+1:h}\,v_j\,(k_j^\top q_t),
    \label{eq:app:decay-output}
\end{equation}
and at a flush the new checkpoint is
$\gamma_L S_{\mathrm{ckpt}}+\sum_{j=1}^{L}\gamma_{j+1:L}v_jk_j^\top$
(Equation~\ref{eq:replay}). Both checkpoint terms are scaled by a cumulative
decay that is data dependent and differs across heads. A head with fast
decay reaches a negligible $\gamma_h$ within a few steps of a window, while
a slow-decaying head, which carries information over long horizons, does
not.

Once $\gamma_h$ falls below what the output precision resolves, the
checkpoint term can be dropped for that head. A non-flush step then moves
only the buffer reads and the record write, and a flush no longer depends on
the old checkpoint and becomes a pure write. In the traffic analysis of Appendix~\ref{app:analysis:traffic},
the $L$ checkpoint reads per window disappear for such a head and leave one
checkpoint write plus the buffer traffic, so the state traffic of that head
falls well below half of conventional decoding. Since the checkpoint term is
scaled by $\gamma_h$, a head whose decay is small but not negligible could
read a coarser copy of the checkpoint instead of skipping it: the
low-precision read checkpoint of
Section~\ref{sec:replayssm-low-precision} already provides such a copy, and
its rounding error enters the output scaled by $\gamma_h$.

\section{Adoption}
\label{app:adoption}

ReplaySSM is adopted in large-scale production models and open-source
inference software. GLM-5.3-Flash uses ReplaySSM in its production
serving stack \citep{zai2026glm53flash}, and the Z.ai infrastructure
report \citep{zai2026glminfra} describes adding it to the KDA decode
kernel. Kimi-K3 \citep{team2026kimi} caches the SSM inputs of draft
tokens for verification in its KDA layers and cites ReplaySSM as a
concurrent approach. ReplaySSM is also integrated into the open-source
serving systems TensorRT-LLM, vLLM, and SGLang and into the kernel
library FlashInfer.

The authors of this paper contributed the blog post
\citep{daolab2026replayssm} and implementations in
TensorRT-LLM,\footnote{\url{https://github.com/NVIDIA/TensorRT-LLM/pull/14203}}
vLLM,\footnote{\raggedright
\url{https://github.com/vllm-project/vllm/pull/48018},
\url{https://github.com/vllm-project/vllm/pull/52506}, and
\url{https://github.com/vllm-project/vllm/pull/52928}.}
and FlashInfer,\footnote{\raggedright
\url{https://github.com/flashinfer-ai/flashinfer/pull/3975} and
\url{https://github.com/flashinfer-ai/flashinfer/pull/4081}.}
covering standard and speculative decoding for Mamba-2 and GDN.

\end{document}